\documentclass[lettersize,journal]{IEEEtran}
\usepackage{amsmath,amsfonts}
\usepackage{subfigure}
\usepackage{algorithmic}
\usepackage{algorithm}
\usepackage{array}
\usepackage{textcomp}
\usepackage{stfloats}
\usepackage{url}
\usepackage{verbatim}
\usepackage{graphicx}
\usepackage{cite}
\usepackage{tcolorbox}
\usepackage{float}
\usepackage{multirow}
\usepackage{booktabs}
\usepackage{tikz}
\usepackage{caption}    
\usepackage{subcaption} 
\usepackage{float}  
\usepackage{colortbl}
\usepackage {hyperref}
\definecolor{wls}{RGB}{47,82,143} 
\usepackage{arydshln}
\begin{document}

\title{FlexiCrackNet: A Flexible Pipeline for Enhanced Crack Segmentation with General Features Transfered from SAM}

\author{Xinlong Wan, Xiaoyan Jiang$^{*}$, Guangsheng Luo, 
Ferdous Sohel,
Jenqneng Hwang,~\IEEEmembership{Life Fellow,~IEEE} and Michael S. Lew
\thanks{
$^{*}$Corresponding author(e-mail: xiaoyan.jiang@sues.edu.cn)

Xiaoyan Jiang is with  the School of Electronic and Electronics Engineering, Shanghai University of Engineering Science, Shanghai 201620, China, and also with Leiden University, Leiden, EZ 2311, The Netherlands (e-mail:  xiaoyan.jiang@sues.edu.cn, x.jiang@liacs.leidenuniv.nl).

Xinlong Wan and
Guangsheng Luo are with
 the School of Electronic and Electronics Engineering, Shanghai University of Engineering Science, Shanghai 201620, China (e-mail: xinlongwan1217@163.com).

 Ferdous Sohel is with the School of Information Technology, Murdoch University, Murdoch, WA 6150, Australia (e-mail: f.sohel@murdoch.edu.au).

Jenq-Neng Hwang is with the Department of Electrical Engineering, University of Washington, Seattle, WA 98105 USA (e-mail: hwang@uw.edu).

Michael S. Lew is with the Leiden Institute of Advanced Computer
Science, Leiden University, Leiden, EZ 2311, The Netherlands (e-mail: m.s.lew@liacs.leidenuniv.nl).
 }
}

\maketitle

\begin{abstract}
Automatic crack segmentation is a cornerstone technology for intelligent visual perception modules in road safety maintenance and structural integrity systems. Existing deep learning models and ``pre-training + fine-tuning'' paradigms often face challenges of limited adaptability in resource-constrained environments and inadequate scalability across diverse data domains.
To overcome these limitations, we propose FlexiCrackNet, a novel pipeline that seamlessly integrates traditional deep learning paradigms with the strengths of large-scale pre-trained models. 
At its core, FlexiCrackNet employs an encoder-decoder architecture to extract task-specific features.
The lightweight EdgeSAM’s CNN-based encoder is exclusively used as a generic feature extractor, decoupled from the fixed input size requirements of EdgeSAM.
To harmonize general and domain-specific features, we introduce the information-Interaction gated attention mechanism (IGAM), which adaptively fuses multi-level features to enhance segmentation performance while mitigating irrelevant noise. 
This design enables the efficient transfer of general knowledge to crack segmentation tasks while ensuring adaptability to diverse input resolutions and resource-constrained environments.
Experiments show that FlexiCrackNet outperforms state-of-the-art methods, excels in zero-shot generalization, computational efficiency, and segmentation robustness under challenging scenarios such as blurry inputs, complex backgrounds, and visually ambiguous artifacts. 
These advancements underscore the potential of FlexiCrackNet for real-world applications in automated crack detection and comprehensive structural health monitoring systems.

\end{abstract}


\section{Introduction}
\IEEEPARstart{C}{racking} is a prevalent and detrimental type of damage commonly observed on road and pavement surfaces, significantly affecting structural integrity \cite{yang2024automation}\cite{song2024universal}. 
Regular crack detection is crucial for road health assessment and accurate damage prediction, enabling timely maintenance that is vital for ensuring road safety and extending pavement lifespan. 
However, traditional manual inspection methods rely heavily on experienced professionals, making the process labor-intensive, time-consuming, and potentially hazardous \cite{ai2023computer}.
Research on automatic crack segmentation methods can be broadly categorized into traditional image processing-based approaches and deep learning-based techniques. 
Most traditional methods struggle to handle complex crack image backgrounds, often failing to capture intricate spatial details and exhibiting high susceptibility to noise. These limitations significantly undermine their effectiveness and reliability, especially in heterogeneous or noisy environments.

\begin{figure}[!t]
    \centering
    \includegraphics[width=6.9cm,height=2.0cm]{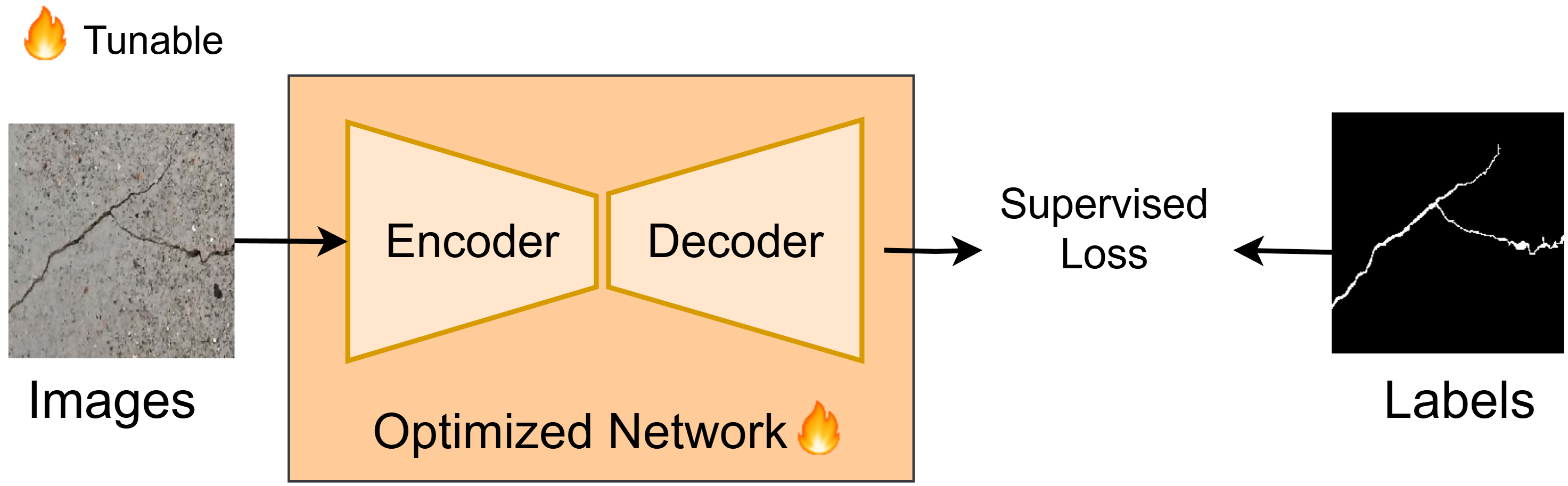} 
    % \vspace{1.5mm}

    \begin{center}
        {\fontsize{9pt}{1pt}\selectfont (a) Paradigm of deep learning-based}
    \end{center}

 \vspace*{-0.3cm}
 
{\color{wls}\rule[0mm]{.85\linewidth}{0.02cm}}
 \vspace*{0.1cm}

    \includegraphics[width=6.9cm,height=2.6cm]{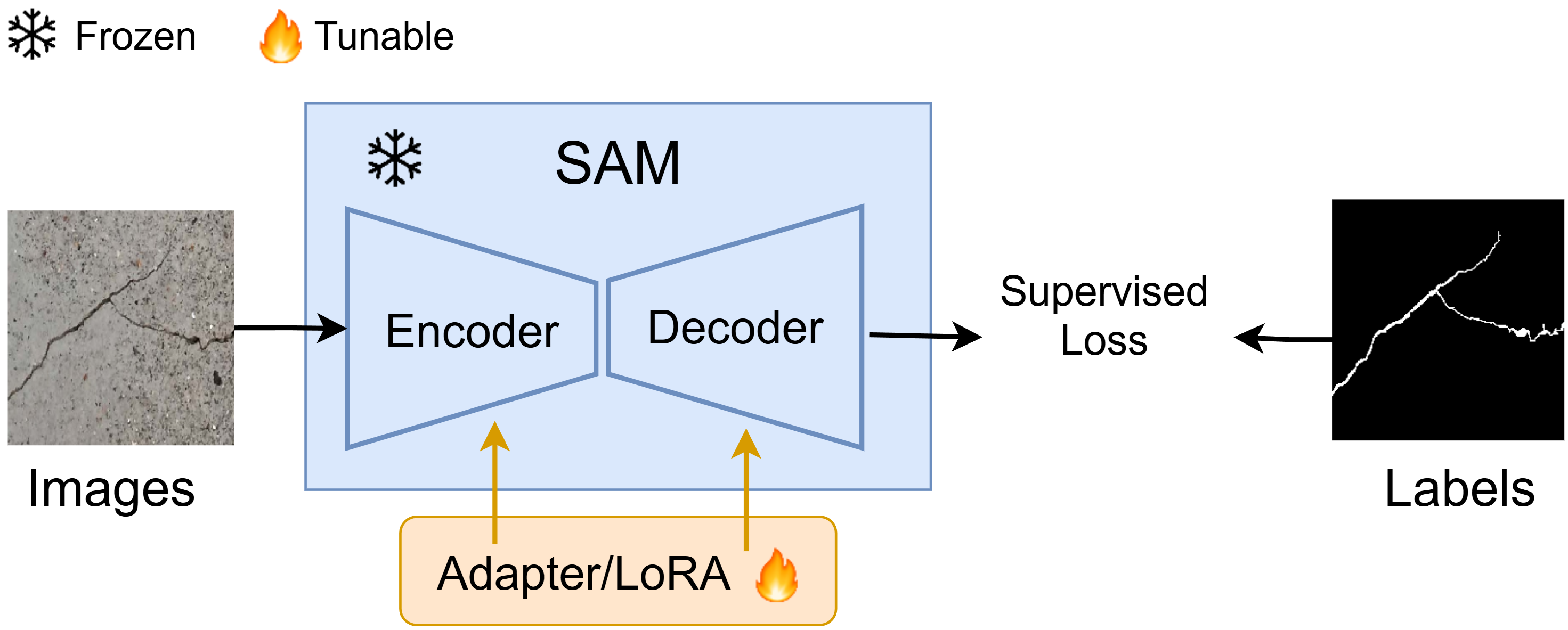} 
    \begin{center}
        {\fontsize{9pt}{1pt}\selectfont (b) The SAM``pre-training + fine-tuning'' Paradigm}
    \end{center}

     \vspace*{-0.3cm}
 
{\color{wls}\rule[0mm]{.85\linewidth}{0.02cm}}
 \vspace*{0.1cm}

    \includegraphics[width=7.0cm,height=3.8cm]{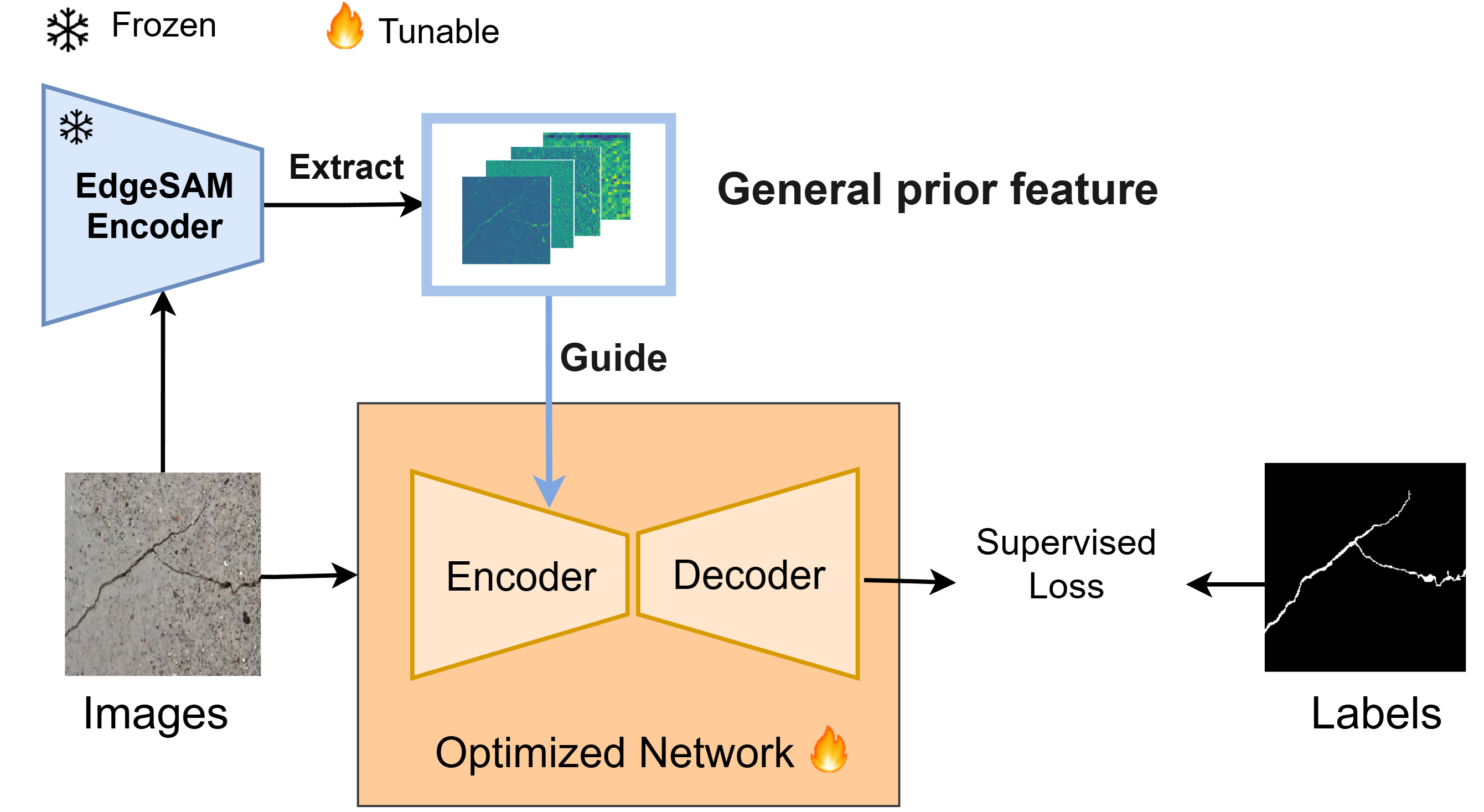} 
    \begin{center}
        {\fontsize{9pt}{1pt}\selectfont (c) The proposed FlexiCrackNet paradigm.}
    \end{center}

    \caption{Crack segmentation training paradigms. (a) Traditional supervised learning allows flexibility in image sizes and model architectures but suffers from limited generalization capability. (b) The ``pre-training + fine-tuning'' paradigm, as exemplified by SAM, offers improved generalization but is constrained by fixed image resolutions and limited architectural adaptability. (c) The proposed FlexiCrackNet supports customizable image sizes and model architectures while significantly enhancing generalization ability.
    }
    \label{Fig1}
\end{figure}

Deep learning approaches, particularly convolutional neural networks (CNNs)\cite{10198487}\cite{10185116}, transformers\cite{10227346}\cite{zhou2023hybrid}, and their combinations \cite{10268450}\cite{10360860} have demonstrated superior performance on crack segmentation. 
These methods can directly learn complex spatial patterns and feature representations from labeled data. 
CNNs excel at capturing local patterns across multiple scales\cite{xie2024ghostformer}, while transformer models leverage self-attention mechanisms to model long-range dependencies\cite{10124821}, thereby enabling more detailed and comprehensive structural representations. 
This ability to capture intricate details significantly enhances segmentation accuracy, establishing deep learning as the dominant paradigm in crack segmentation\cite{gong2024state}. 
The training paradigm for deep learning methods, illustrated in Fig. \ref{Fig1} (a), relies on labeled datasets and model training. However, a persistent challenge in crack segmentation is the limited generalization capability of these models. Due to the relatively small size of available crack segmentation datasets, models often overfit to the training data, leading to diminished performance on test sets and under diverse real-world road conditions \cite{chen2023automatic}. 
Although strategies such as data augmentation, regularization, and hyperparameter tuning have been employed to address overfitting \cite{alomar2023data}\cite{li2023understanding}, these methods frequently require substantial computational resources and extensive experimentation, which can limit their practical scalability and effectiveness.

Recently, large-scale foundational models have demonstrated remarkable performance across a wide range of tasks. Models such as GPT \cite{dale2021gpt}, CLIP \cite{radford2021learning}, and the Segment Anything model (SAM) \cite{kirillov2023segment} exhibit high generalization capabilities in both language and vision domains. 
Among these, SAM distinguishes itself as a versatile image segmentation model, leveraging robust semantic understanding and spatial feature extraction abilities gained from extensive training on large-scale visual datasets. The adoption of such foundational models for downstream tasks offers a promising solution to challenges posed by limited training data and enhances the generalization capacity of feature extraction \cite{10267997}\cite{chen2023sam}.
Subsequently, the ``pre-training + fine-tuning'' paradigm has emerged as a pivotal strategy in recent research \cite{mazurowski2023segment}\cite{ke2024segment}, particularly for adapting models like SAM to downstream tasks with limited annotated datasets. 
Given the high cost and labor-intensive nature of pixel-level annotations in crack segmentation, this pipeline is especially valuable for improving task performance. Studies have demonstrated that fine-tuning SAM for crack segmentation achieves superior results compared to traditional methods \cite{zhou2024teaching}. The process involves freezing the core parameters of the pre-trained model while incorporating smaller, task-specific modules for adaptation, enabling efficient fine-tuning without the need to retrain the entire network, as illustrated in Fig. \ref{Fig1} (b).

Despite the impressive performance of the SAM ``pre-training + fine-tuning'' paradigm in crack segmentation tasks, it faces notable limitations in scalability and flexibility. For instance, fine-tuning SAM necessitates resizing images to 1024$\times$1024, which restricts its adaptability to downstream datasets with diverse input sizes.  
Additionally, the high memory consumption associated with processing such large images hinders the feasibility of deploying the model on resource-constrained mobile platforms, reducing its practicality for real-world applications in such settings.

To overcome these challenges, we introduce FlexiCrackNet, a new pipeline that redefines how general-purpose segmentation models are utilized for crack detection. 
To ensure computational efficiency, we adopt EdgeSAM \cite{zhou2023edgesam}, a lightweight version of SAM specifically designed for resource-constrained edge devices. 
EdgeSAM employs a CNN-based encoder paired with a transformer-based decoder, limiting its input resolution to 1024$\times$1024. 
An overview of EdgeSAM is given in Section \ref{subsec:A}.
In our proposed paradigm, we decouple this limitation by using EdgeSAM’s CNN-based encoder exclusively as a generic feature extractor. 
This allows FlexiCrackNet to process input images of arbitrary sizes, significantly enhancing flexibility. As illustrated in Fig. \ref{Fig1} (c), the extracted multi-level features, spanning low-level textures to high-level semantic representations, are then utilized to guide the training of a custom crack segmentation model tailored to the unique characteristics of crack-specific tasks.

This new paradigm integrates general visual representations with crack-specific insights in a way that optimizes both generalization and efficiency. Experimental results with zero-shot evaluation show that FlexiCrackNet substantially improves crack segmentation accuracy while maintaining minimal additional inference costs. By shifting from rigid frameworks to a more flexible and task-adaptive paradigm, FlexiCrackNet establishes a foundation for efficient, scalable, and practical solutions to crack segmentation challenges in real-world environments.
The main contributions are summarized as follows:

(1) We introduce FlexiCrackNet, a novel pipeline that integrates the flexibility of traditional deep learning paradigms with the strengths of the SAM-based ``pre-training + fine-tuning'' approach, effectively overcoming its limitations in crack segmentation tasks. This pipeline enables the efficient transfer of general knowledge from pre-trained models while ensuring adaptability to diverse datasets and resource-constrained environments.

(2) We identify the challenge of noise introduced by non-crack-related features in multi-level generic representations, which capture both low-level texture details and high-level semantic structures. To address this, we propose an information-interaction gated attention mechanism, which selectively fuses multi-level features with crack-specific representations, enhancing segmentation accuracy while minimizing irrelevant noise.

(3) Experimental results demonstrate that FlexiCrackNet surpasses state-of-the-art methods in both supervised and zero-shot scenarios, establishing a new benchmark for crack segmentation performance.

\section{RELATED WORK}
\subsection{Deep Supervision in Crack Segmentation}
Recent advancements in deep learning-based crack segmentation have centered on achieving precise, fine-grained pixel-level detection across various datasets \cite{zhang2023ecsnet, quan2023crackvit, al2023hybrid}. Well-established CNN architectures, including fully convolutional networks (FCN) \cite{long2015fully}, U-Net \cite{ronneberger2015u}, and SegNet \cite{badrinarayanan2017segnet}, have been widely adopted due to their ability to handle variable input image sizes and capture intricate spatial details essential for accurate crack detection\cite{zheng2024tv}.
FCN architectures employ skip connections to integrate features across multiple network layers, allowing a seamless fusion of high-level semantic information with low-level spatial details during upsampling. This design effectively preserves both global context and local details, ensuring robust segmentation performance critical for delineating cracks at the pixel level.

U-Net adopts a symmetric encoder-decoder design. The encoder progressively reduces spatial resolution while extracting high-level features through convolution and pooling layers, whereas the decoder restores spatial resolution via upsampling. Skip connections between corresponding encoder and decoder layers allow the retention of detailed spatial features, enhancing segmentation accuracy. For instance, Fan et al. \cite{fan2021nested} demonstrated the efficacy of U-Net in capturing detailed crack features by adopting an encoder-decoder structure with skip connections to fuse feature maps from various stages. SegNet employs an encoder-decoder architecture similar to U-Net but enhances efficiency by storing max-pooling indices for accurate spatial reconstruction during decoding.

Both U-Net and SegNet exemplify the effectiveness of encoder-decoder architectures in achieving robust performance for image segmentation by preserving spatial details and reconstructing meaningful features. Encoder-decoder frameworks, such as U-Net, FCN, and SegNet, have become the backbone of crack segmentation due to their ability to extract hierarchical feature representations while maintaining spatial coherence. However, despite their strengths, these architectures face challenges in feature fusion and representation learning when addressing complex or low-contrast cracks. These limitations underscore the need for further advancements to improve their robustness and adaptability in diverse and challenging environments.

In recent years, transformer-based semantic segmentation has garnered significant attention due to their superior capability for global feature extraction and context integration \cite{ranftl2021vision}. A notable example is SegFormer \cite{xie2021segformer}, integrating transformer mechanisms into the semantic segmentation process. SegFormer employs multi-head self-attention mechanisms and hierarchical feature representations to effectively capture both global contextual information and local details. Compared to traditional CNN-based approaches, transformer architectures demonstrate substantial improvements in segmentation performance by providing enhanced contextual understanding \cite{shamsabadi2022vision}\cite{tao2023convolutional}.

However, due to the limited size of crack segmentation datasets, models are highly susceptible to overfitting, often memorizing specific crack features from the training data rather than learning generalized patterns. Consequently, their performance during inference is frequently suboptimal, particularly when applied to unseen data with varying characteristics. Addressing this limitation requires innovative approaches to enhance the generalization capabilities of crack segmentation models and mitigate overfitting. 
Unlike traditional methods that fuse features learned directly from the input image, our pipeline integrates generalized features from a pre-trained segmentation model into the fusion process. These broader, more transferable features promote flexible learning, thereby reducing the risk of overfitting. By incorporating these generalized features, our approach enhances the model's ability to generalize across diverse datasets. This method improves the downstream model's ability to effectively identify cracks, ensuring better performance across varied datasets and real-world scenarios.

\subsection{Transferring Knowledge of Vision Foundation Models}
Recently, the release of SAM \cite{kirillov2023segment} has introduced a new paradigm in the field of image segmentation. 
Proposed by Meta AI, SAM is a general-purpose segmentation model designed to achieve the ambitious goal of "segmenting anything." 
Pre-trained on an unprecedented scale—1.1 billion masks across 11 million images—SAM represents a significant breakthrough in computer vision by demonstrating robust zero-shot performance across diverse semantic segmentation tasks. 
Despite its remarkable generalization capabilities, SAM's tendency to generate masks for all distinguishable instances in an image limits its applicability in specialized tasks, such as camouflaged object detection, medical image segmentation, and crack detection, where precision and domain-specific adaptations are critical.

To address these limitations, recent research has increasingly adopted parameter-efficient fine-tuning (PEFT) techniques to adapt SAM for specific downstream applications. 
PEFT methods, such as adapters\cite{houlsby2019parameter}, introduce lightweight trainable parameters into pre-trained models, allowing efficient domain adaptation while preserving the original model’s knowledge and avoiding catastrophic forgetting. For instance, Chen et al. proposed SAM-Adapter, which demonstrated superior performance in camouflaged object detection, shadow detection, and polyp segmentation\cite{chen2023sam}. 

In the domain of crack segmentation,
Zhou et al. adapted SAM for road crack segmentation by incorporating Adapter modules into its image encoder, enabling the model to learn domain-specific knowledge relevant to crack detection\cite{zhou2024teaching}. These studies collectively underscore SAM’s transformative potential in image segmentation while highlighting the importance of targeted adaptations to meet the demands of specialized tasks.

However, challenges persist, particularly regarding the high computational cost and memory demands of SAM during training and inference. SAM’s architecture, which includes multiple transformer layers, requires significant computational resources, especially when processing images at the fixed 1024$\times$1024 resolution dictated by the pre-trained model. This limitation poses significant obstacles to deploying SAM on resource-constrained devices, such as those used for mobile-based crack detection. 
To address this, we propose a flexible pipeline that allows customization of both input image size and model architecture, enabling adaptation to available computational resources while effectively leveraging the general knowledge embedded in the foundational model.

\section{PROPOSED METHOD}
In Section \ref{subsec:A}, we evaluate the feature extraction capabilities of the universal feature extractor, EdgeSAM's Encoder.
The proposed pipeline is illustrated in Section \ref{subsec:B}, showing how FlexiCrackNet incorporates general prior knowledge with domain-specific features in an efficient way to realize an enhanced crack segmentation.
Section \ref{subsec:C} shows the loss function used for FlexiCrackNet. 

\subsection{EdgeSAM Encoder}
\label{subsec:A}
We begin by reviewing EdgeSAM, the universal segmentation model that forms the foundation of our approach. EdgeSAM is a streamlined, accelerated variant of SAM, specifically optimized for efficient operation on edge devices. In EdgeSAM, the original ViT-based SAM image encoder is distilled into a CNN-based architecture, making it better suited for edge deployment. Compared to the original SAM, EdgeSAM achieves an impressive 37$\times$ speed improvement and is the first SAM variant capable of running at over 30 FPS on an iPhone 14.

\begin{figure}[!t]
	\centering
	\begin{minipage}{0.49\linewidth}
		\centering
		\includegraphics[width=.7\linewidth]{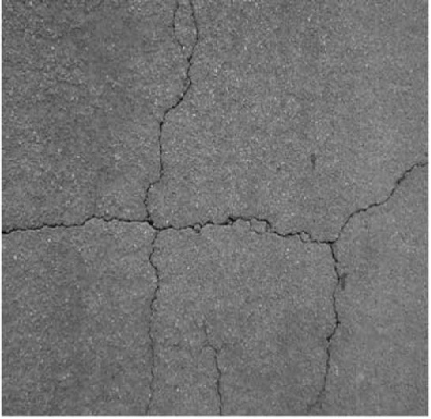}\\
        (a) Original image
	\end{minipage}
    	\begin{minipage}{0.49\linewidth}
		\centering
        \resizebox{\linewidth}{!}{
        \renewcommand{\arraystretch}{0.4}  % Increase line spacing by 50%
        \begin{tabular}{p{2.4cm}p{2.4cm}p{2.4cm}}
         \includegraphics[width=2.8cm,height=2.8cm]{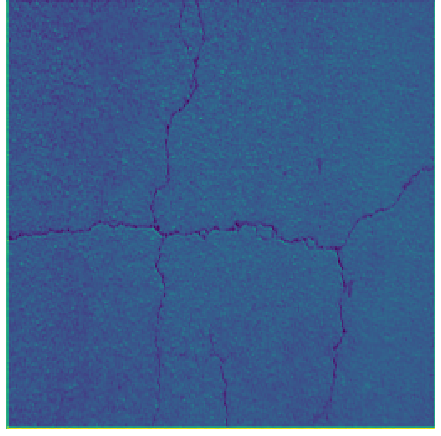}
         &\includegraphics[width=2.8cm,height=2.8cm]{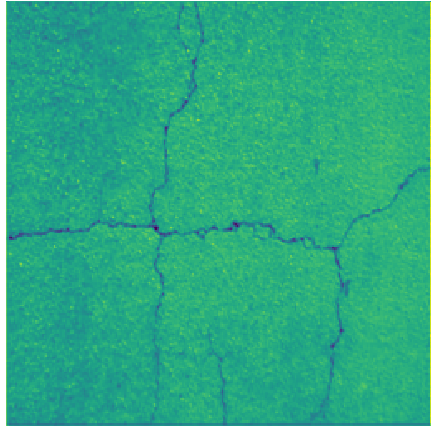}&\includegraphics[width=2.8cm,height=2.8cm]{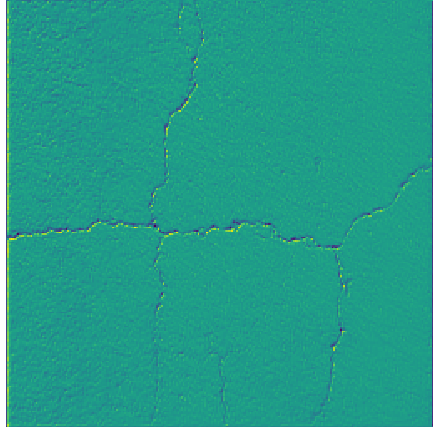}\\
          \includegraphics[width=2.8cm,height=2.8cm]{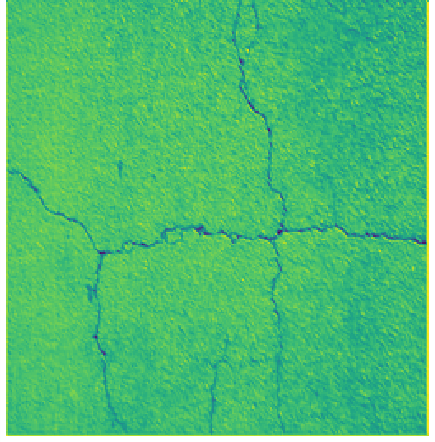}&\includegraphics[width=2.8cm,height=2.8cm]{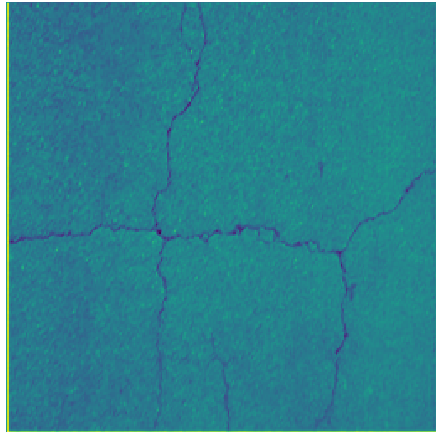}&\includegraphics[width=2.8cm,height=2.8cm]{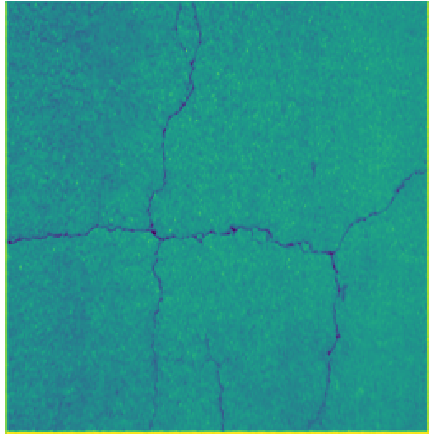}\\
           \includegraphics[width=2.8cm,height=2.8cm]{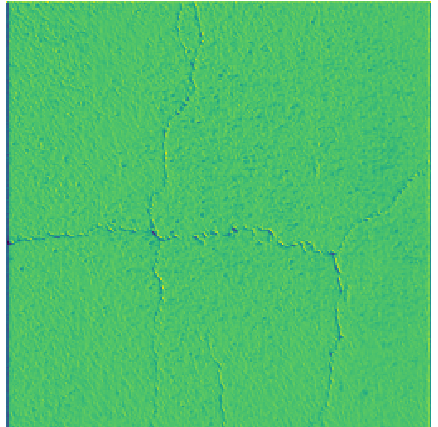}&\includegraphics[width=2.8cm,height=2.8cm]{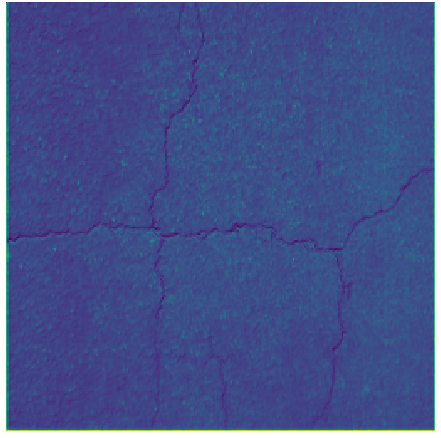}&\includegraphics[width=2.8cm,height=2.8cm]{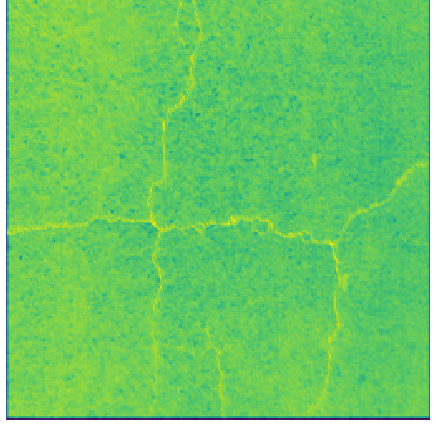}\\
        \end{tabular}}\\
        \vspace{1mm}
        (b) Stage1
	\end{minipage}
 \vspace{1.5mm}
 
      	\begin{minipage}{0.49\linewidth}
		\centering
        \resizebox{\linewidth}{!}{
        \renewcommand{\arraystretch}{0.4}  % Increase line spacing by 50%
        \begin{tabular}{p{2.4cm}p{2.4cm}p{2.4cm}}
         \includegraphics[width=2.8cm,height=2.8cm]{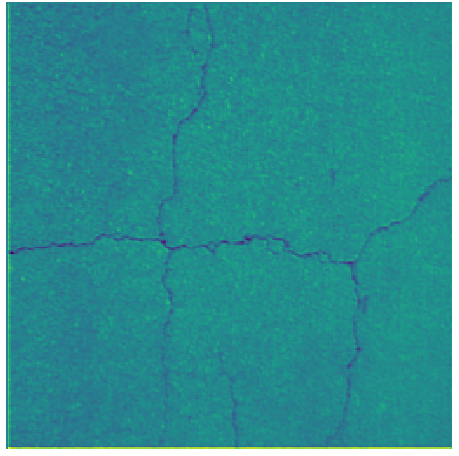}&\includegraphics[width=2.8cm,height=2.8cm]{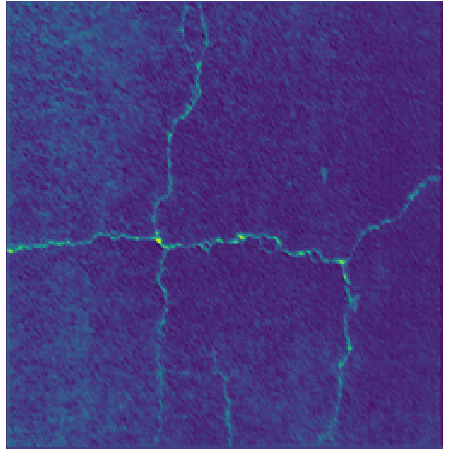}&\includegraphics[width=2.8cm,height=2.8cm]{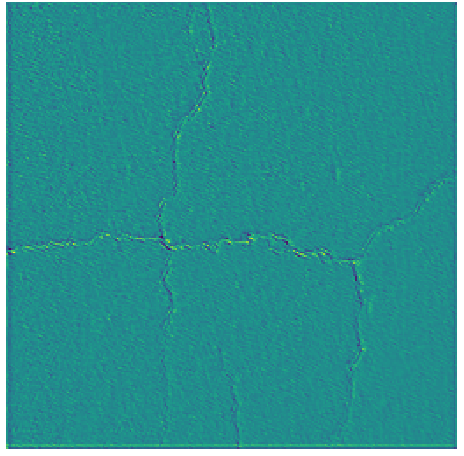}\\
          \includegraphics[width=2.8cm,height=2.8cm]{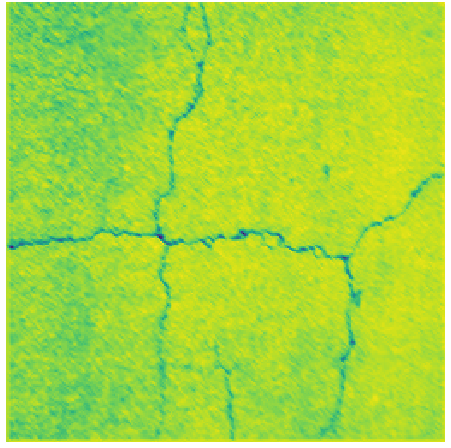}&\includegraphics[width=2.8cm,height=2.8cm]{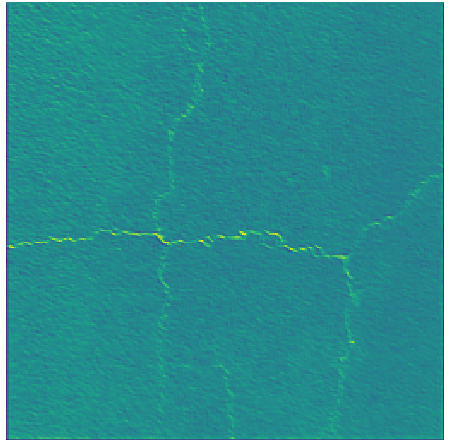}&\includegraphics[width=2.8cm,height=2.8cm]{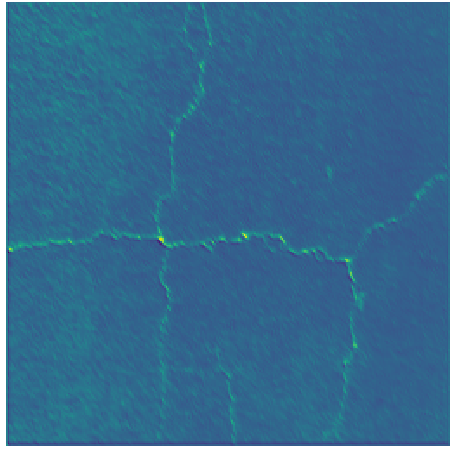}\\
           \includegraphics[width=2.8cm,height=2.8cm]{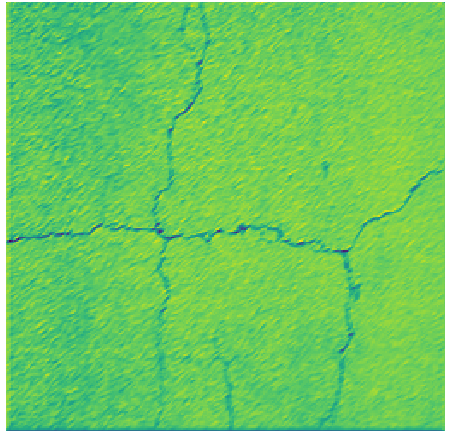}&\includegraphics[width=2.8cm,height=2.8cm]{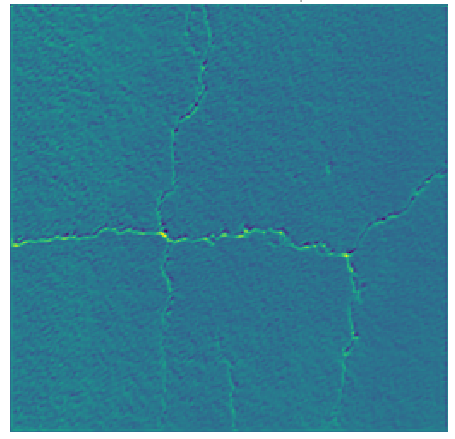}&\includegraphics[width=2.8cm,height=2.8cm]{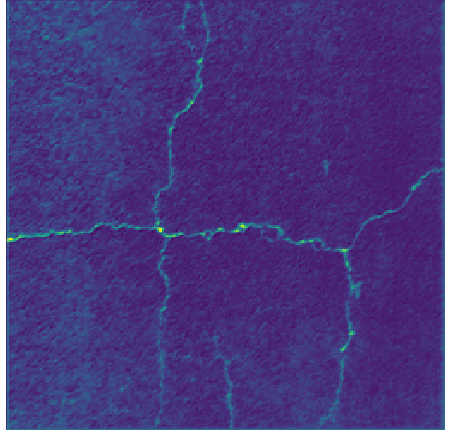}\\
        \end{tabular}}\\
         \vspace{1mm}
        (c) Stage2
	\end{minipage}
      	\begin{minipage}{0.49\linewidth}
		\centering
        \resizebox{\linewidth}{!}{
        \renewcommand{\arraystretch}{0.4}
        \begin{tabular}{p{2.4cm}p{2.4cm}p{2.4cm}}
         \includegraphics[width=2.8cm,height=2.8cm]{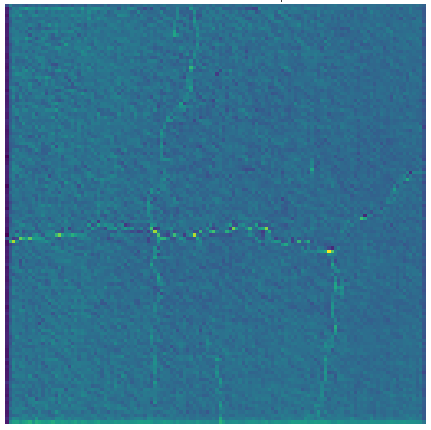}&\includegraphics[width=2.8cm,height=2.8cm]{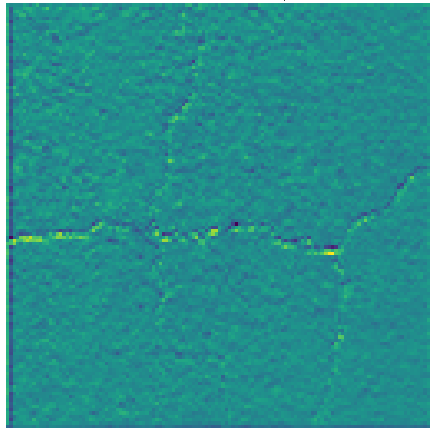}&\includegraphics[width=2.8cm,height=2.8cm]{figure/stage1/3.jpg}\\
          \includegraphics[width=2.8cm,height=2.8cm]{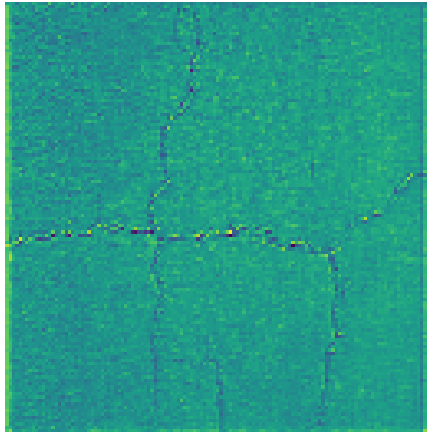}&\includegraphics[width=2.8cm,height=2.8cm]{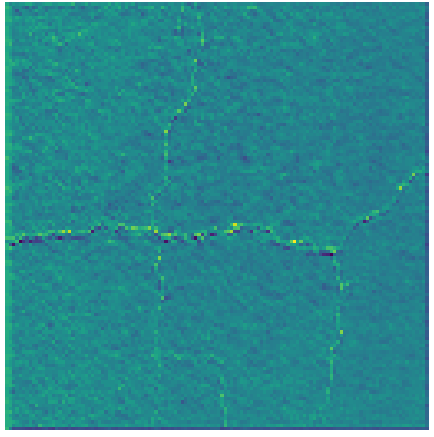}&\includegraphics[width=2.8cm,height=2.8cm]{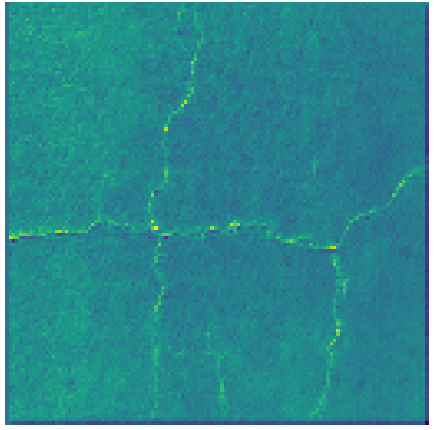}\\
           \includegraphics[width=2.8cm,height=2.8cm]{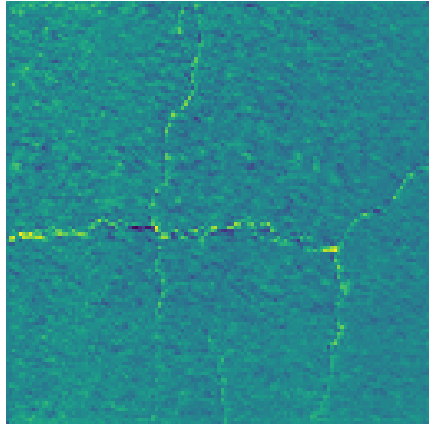}&\includegraphics[width=2.8cm,height=2.8cm]{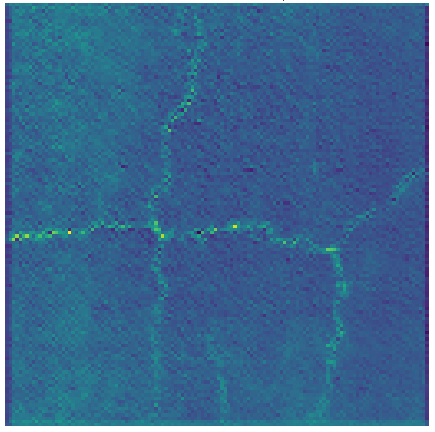}&\includegraphics[width=2.8cm,height=2.8cm]{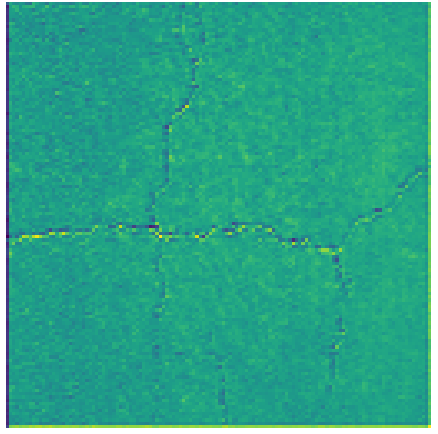}\\
        \end{tabular}}\\
         \vspace{1mm}
        (d) Stage3
	\end{minipage}
 \vspace{1.5mm}
 
      	\begin{minipage}{0.49\linewidth}
		\centering
        \resizebox{\linewidth}{!}{
        \renewcommand{\arraystretch}{0.4}
        \begin{tabular}{p{2.4cm}p{2.4cm}p{2.4cm}}
         \includegraphics[width=2.8cm,height=2.8cm]{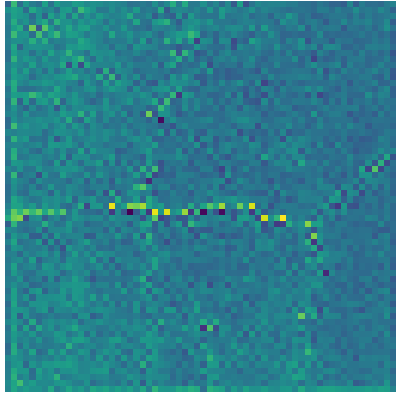}&\includegraphics[width=2.8cm,height=2.8cm]{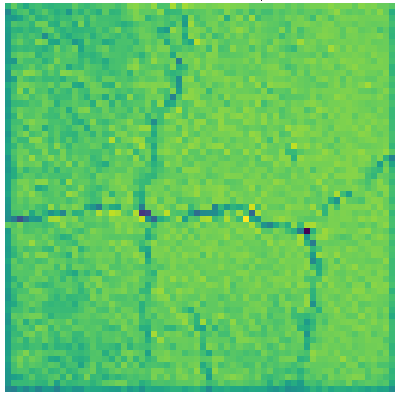}&\includegraphics[width=2.8cm,height=2.8cm]{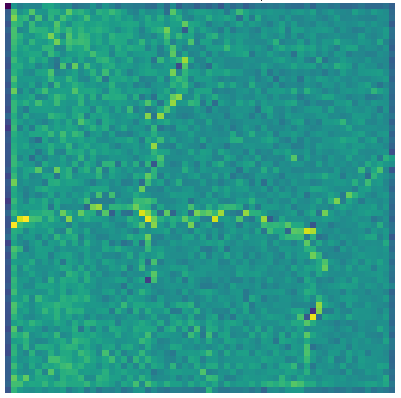}\\
          \includegraphics[width=2.8cm,height=2.8cm]{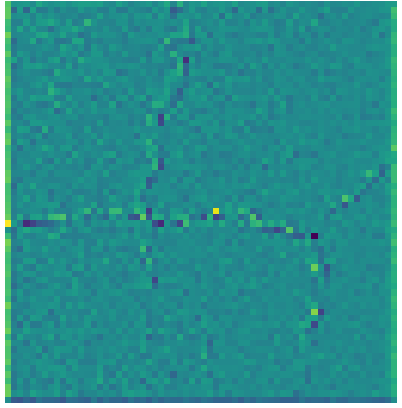}&\includegraphics[width=2.8cm,height=2.8cm]{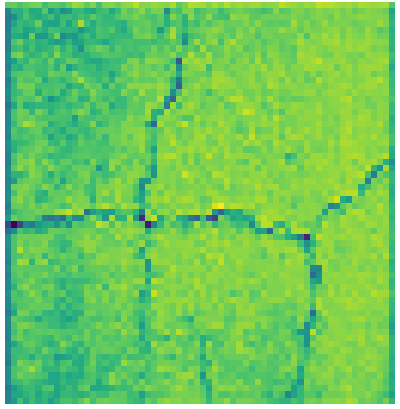}&\includegraphics[width=2.8cm,height=2.8cm]{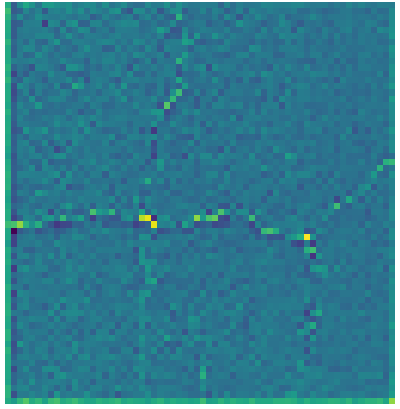}\\
           \includegraphics[width=2.8cm,height=2.8cm]{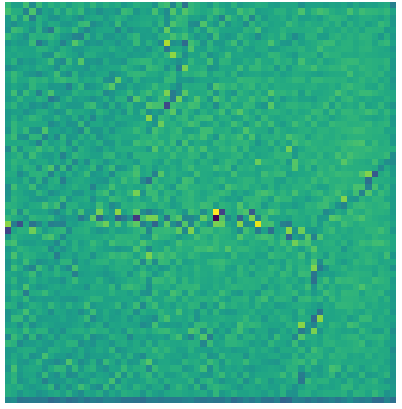}&\includegraphics[width=2.8cm,height=2.8cm]{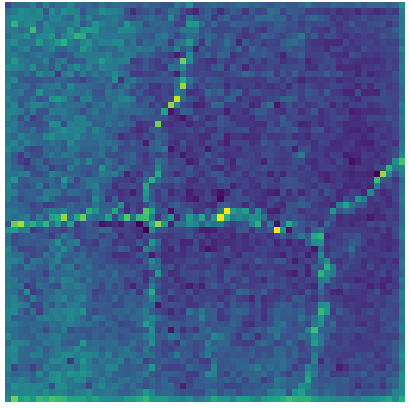}&\includegraphics[width=2.8cm,height=2.8cm]{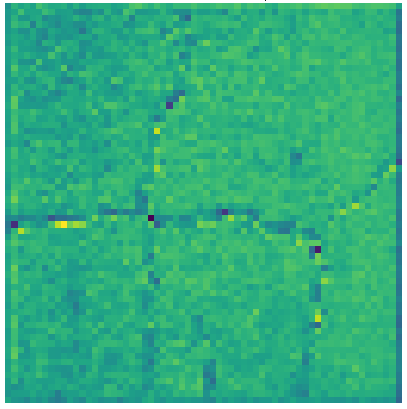}\\
        \end{tabular}}\\
         \vspace{1mm}
        (e) Stage4
	\end{minipage}
      	\begin{minipage}{0.49\linewidth}
		\centering
        \resizebox{\linewidth}{!}{
         \renewcommand{\arraystretch}{0.2}
        \begin{tabular}{p{2.4cm}p{2.4cm}p{2.4cm}}
         \includegraphics[width=2.8cm,height=2.8cm]{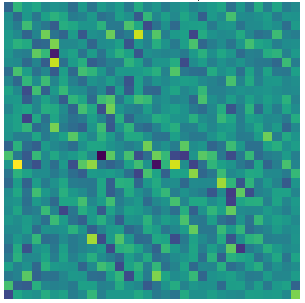}&\includegraphics[width=2.8cm,height=2.8cm]{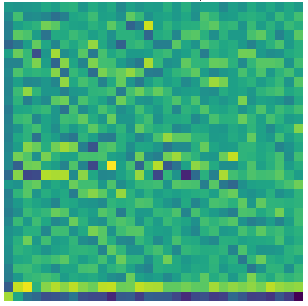}&\includegraphics[width=2.8cm,height=2.8cm]{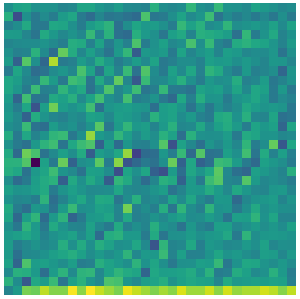}\\
          \includegraphics[width=2.8cm,height=2.8cm]{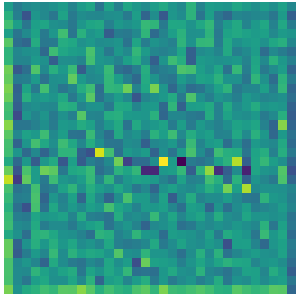}&\includegraphics[width=2.8cm,height=2.8cm]{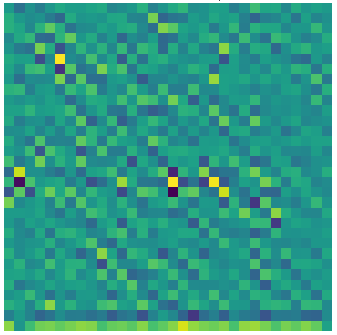}&\includegraphics[width=2.8cm,height=2.8cm]{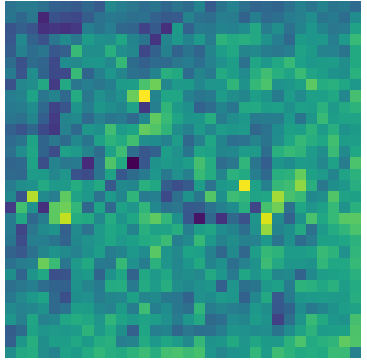}\\
           \includegraphics[width=2.8cm,height=2.8cm]{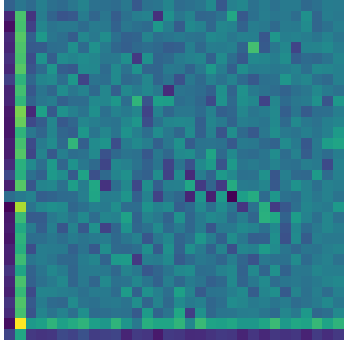}&\includegraphics[width=2.8cm,height=2.8cm]{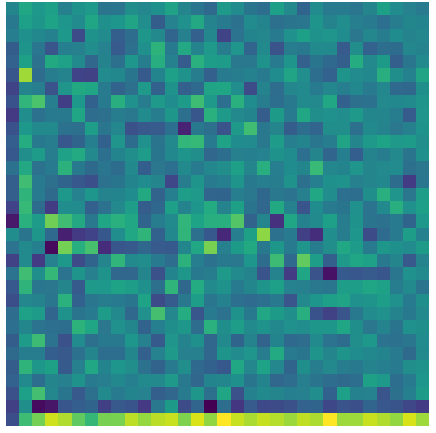}&\includegraphics[width=2.8cm,height=2.8cm]{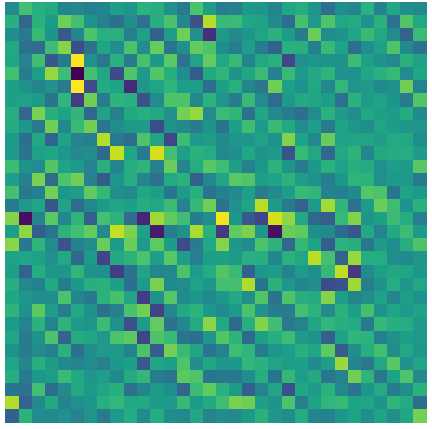}\\
        \end{tabular}}\\
         \vspace{1mm}
        (f) Stage5
	\end{minipage}

    \caption{The first nine feature maps of the five stages in the EdgeSAM encoder.}
    \label{Fig2}
\end{figure}

EdgeSAM employs a CNN-based architecture for its encoder and transformer layers for its decoder. The transformer-based decoder requires consistency with the pre-trained input size, which is fixed at a resolution of 1024$\times$1024. As a result, the pre-trained model is constrained to process images of this specific size. To address the need for flexible and customized image sizes in crack segmentation tasks, we exclusively utilize EdgeSAM's CNN-based encoder as a universal feature extractor, enabling adaptability while maintaining efficient feature representation.
To assess the effectiveness of the universal features extracted by EdgeSAM, we evaluate its performance in zero-shot crack image feature extraction, as illustrated in Fig. \ref{Fig2}. The figure depicts feature maps generated from input images through the EdgeSAM encoder, which has a total of four downsampling layers. The input image passes through a total of four downsampling layers, including the stage before the first downsampling, resulting in five stages of feature maps. These feature maps are visualized to show how the features evolve across different resolutions, demonstrating the progression of feature extraction through the five stages.

The figure depicts feature maps generated from input images through the four downsampling layers of the EdgeSAM encoder. The results indicate that, even in a zero-shot setting, the encoder can effectively extract both clear and detailed low-level features—such as intricate crack textures and boundary details—and high-level semantic information, capturing the broader context of the image. 
These features encapsulate fine-grained structural details alongside higher-level representations of the image content, making them essential for tasks like crack segmentation. 

Building on these promising findings, we further investigate how to seamlessly integrate these universal prior features into a robust crack segmentation pipeline.
\subsection{The Proposed Model Architecture}
\label{subsec:B}
 \textbf{Discussion on the encoder-decoder architecture.} 
Encoder-decoder architectures are the dominant paradigm for crack segmentation tasks due to their ability to effectively capture fine-grained features through successive downsampling and upsampling processes. These architectures leverage skip connections to integrate shallow features from early layers with deeper, high-level features, enabling improved detail recovery and spatial consistency. The encoder extracts hierarchical feature representations by progressively reducing spatial resolution, while the decoder restores spatial details through upsampling, reconstructing the image structure with enhanced precision.

In this paper, we adopt the U-Net architecture as the baseline for our encoder-decoder design, optimizing its strengths to develop a novel crack segmentation pipeline that incorporates generalized visual prior features. U-Net is chosen for its proven effectiveness and versatility across diverse segmentation tasks. While other encoder-decoder architectures can theoretically integrate generalized prior knowledge, U-Net’s robust performance and broad applicability make it an ideal choice for this work.

A key advantage of U-Net is its use of skip connections at each stage, directly transferring encoder-extracted features to the corresponding decoder layers. This feature fusion is essential for recovering fine-grained details such as cracks, ensuring high segmentation accuracy. Additionally, U-Net is computationally efficient, with fewer parameters compared to many other architectures, reducing the risk of overfitting when working with limited datasets. These characteristics establish U-Net as a solid foundation for crack segmentation tasks, providing both the feature integration capabilities and efficiency required for our proposed pipeline.

\begin{figure*}[!t]
	\centering
	\includegraphics[width=17.5cm,height=12.2cm]{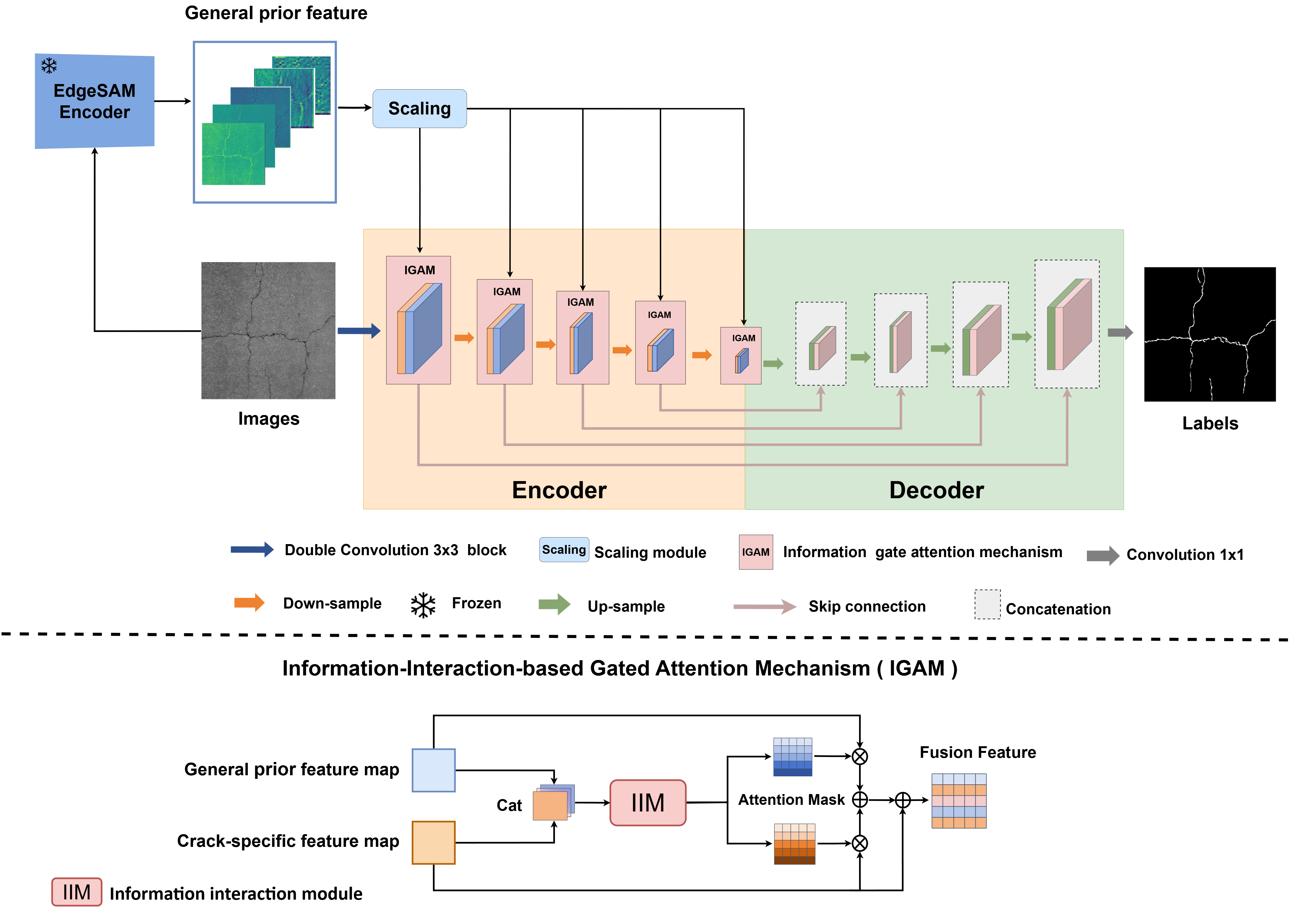}
    \caption{The overall pipeline of FlexiCrackNet. The pipeline adopts an encoder-decoder structure for crack segmentation, where IGAM modules are integrated into the encoder to fuse general semantic and crack-specific features. IIM within IGAM generates crack-specific attention masks, enabling effective feature fusion.}
    \label{Fig3}
\end{figure*}

\textbf{FlexiCrackNet.} 
Fig. \ref{Fig3} illustrates the proposed FlexiCrackNet pipeline, which consists of three key modules: the image encoder, the universal feature extractor (EdgeSAM Encoder), and the image decoder. 
Specifically, the original crack image is fed in parallel into both the image encoder and the EdgeSAM Encoder. The image encoder is responsible for learning semantic feature representations of the cracks during training. 
Meanwhile, EdgeSAM Encoder remains frozen during training and serves to extract multi-level, general-purpose visual features from the raw image. These multi-level features span both low-level spatial details (such as fine-grained edges and textures) and high-level semantic abstractions (such as structural patterns and contextual relationships). By integrating these generic features as multi-level prior cues with the multi-level features learned by the image encoder, the encoder gains richer prior information, which provides the decoder with more valuable input. The encoder of the proposed pipeline consists of five stages. 

During the forward propagation, both the image encoder and the EdgeSAM encoder pass through four downsampling modules, generating feature maps at five distinct stages. At each stage, we align and fuse the feature map generated by the image encoder with the generic feature map produced by the EdgeSAM encoder at the same stage, using the proposed IGAM. During training, only the encoder and decoder weights are updated, ensuring that the segmentation model can effectively learn to adapt to the specific characteristics of crack images while benefiting from the generalized visual priors embedded in EdgeSAM Encoder. This design provides an efficient mechanism for leveraging both domain-specific features and generalized visual knowledge to improve segmentation performance.

\textbf{Scaling module.} 
Before fusing the generalized prior features with the crack-specific features learned by the segmentation model, it is essential to align them to the same scale, both channel and spatial dimensions. This ensures that the features can be effectively combined during the fusion step. The operations performed in the scaling module are as follows:

\begin{equation}
\label{deqn_ex1}
F_ {scaled} = PwConv(Reshape( F_ {general} ))~,
\end{equation}
where $Reshape( \cdot )$ represents reshapes the input features to align with the required spatial dimensions, $F_ {general}$ reshapes the general prior feature, 
and $PwConv$ represents a 1×1 convolution that adjusts the number of channels to match the required channel count. 

\textbf{IGAM.} General prior features may contain irrelevant information and may not align with the model's learned feature space. To address this, we design IGAM, which facilitates feature fusion at each encoder stage. The structure of IGAM is shown in Fig \ref{Fig3}. In IGAM, the general prior feature map and crack-specific feature map are concatenated along the channel dimension, then processed by an information interaction module that generates two attention masks. The first mask highlights meaningful regions in the general prior feature map, while the second enhances relevant details in the crack-specific feature  map. These modulated maps are combined through element-wise addition to produce a fused feature representation. To prevent information dilution, IGAM uses residual connections, directly adding the original crack-specific feature map to the fused representation. This ensures that critical crack-related semantics are preserved throughout the fusion process.

\begin{figure}[!t]
    \centering
    \includegraphics[width=8.3cm,height=3.6cm]{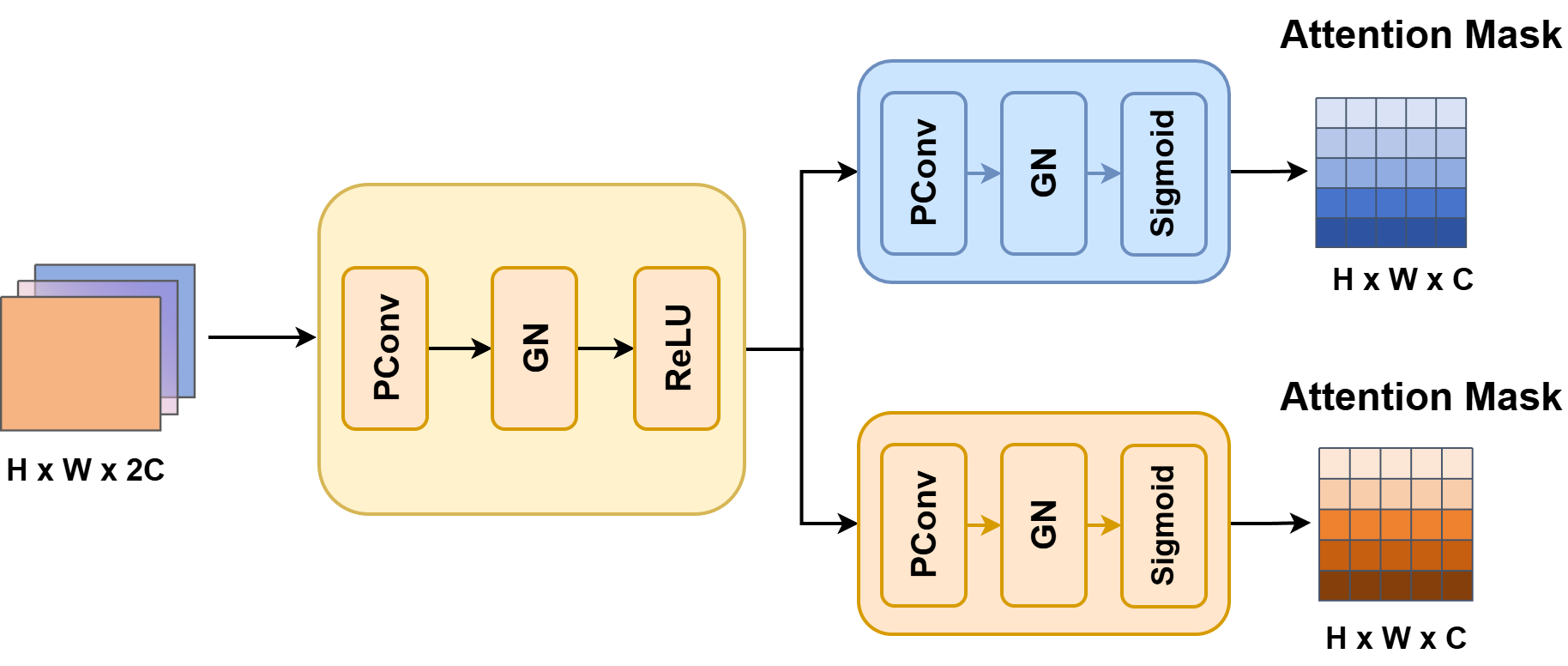}
    \caption{The structure of IIM.
    }
    \label{Fig4}
\end{figure}

\textbf{Information interaction module (IIM).} 
As shown in Fig \ref{Fig4}, the information interaction module processes the concatenated feature map (H×W×2C) to generate two attention masks for general semantic and crack-specific features. First, the concatenated feature map passes through a Pointwise Convolution (PConv) layer to reduce dimensionality while preserving spatial context, followed by Group Normalization (GN) and ReLU activation to promote channel-wise feature interaction. The refined features are then split into two branches, where each branch applies PConv, GN, and Sigmoid activation to produce a normalized attention mask (H×W×C). These masks selectively emphasize task-relevant information, ensuring optimal alignment and feature fusion in later stages.

\subsection{Loss Function}
\label{subsec:C}
To address the class imbalance in crack segmentation and improve model stability during training, we combine Binary Cross-Entropy (BCE) Loss and Dice Loss as the supervised loss function. BCE Loss provides reliable gradients by calculating the loss for each pixel independently, as follows:
\begin{equation}
L_{Bce} =-\frac{1}{N} \sum_{i=1}^{N} y_{i} \log \left(p_{i}\right)+\left(1-y_{i}\right) \log \left(1-p_{i}\right)~,
\end{equation}
where $p_{i}$ and $y_{i}$ are the predicted probability value and the annotated label of the $i^{th}$ pixel, respectively. N is the total number of pixels in the image. 

BCE Loss struggles with class imbalance, particularly for underrepresented classes. To address this, we introduce Dice Loss, which enhances performance by focusing on the overlap between the predicted and ground truth masks:

\begin{equation}
L_{Dice} = 1-\frac{2|X \cap Y|}{|X|+|Y|}~,
\end{equation}
where $X$ and $Y$ represent the predicted and ground truth masks, respectively, and $|X \cap Y|$ the intersection of the two. 

The total loss function is a combination of BCE and Dice Losses:
\begin{equation}
% \label{equ:13}
L_{total} = L_{Bce} +  L_{Dice} .
\end{equation}

\section{EXPERIMENTS}
\subsection{Experimental Configuration }
To ensure a fair and consistent comparison across all methods, we standardize the training configuration for all experiments. Input images are resized to 512$\times$512 pixels, and data augmentation is applied using random rotations and flips to enhance generalization. All models are optimized with the AdamW optimizer, utilizing a fixed loss function to maintain uniformity in the training procedure. The initial learning rate is set to 0.0003 and adjusted dynamically using a cosine decay strategy. Training is conducted with a batch size of 2 over a total of 100 epochs. The experiments are implemented using the PyTorch framework on a system equipped with an Intel Xeon Gold 6226R CPU and a single NVIDIA Tesla A100 GPU, ensuring high computational efficiency.

\subsection{Datasets}
To evaluate the performance of the proposed model, we utilize two publicly available pavement crack datasets: the DeepCrack dataset \cite{liu2019deepcrack}, the Cityforest dataset (CFD) \cite{shi2016automatic} and Crack500 dataset\cite{8694955}. 
To assess the model's generalization capability, training is performed exclusively on the training set of the DeepCrack dataset, while evaluation is conducted on both the DeepCrack test set, the entire CFD dataset and Crack500 dataset.

DeepCrack Dataset: It contains 537 manually annotated RGB images of crack patterns observed on various surfaces, including concrete and stone. The dataset captures a diverse range of crack characteristics, such as very wide cracks and blurred regions, providing a comprehensive basis for model training and evaluation. 
It is divided into a training set with 300 images and a test set containing 237 images, enabling robust performance analysis under diverse conditions.

Cityforest Dataset: The CFD dataset focuses on urban road surface conditions in Beijing, consisting of 118 pavement crack images captured using an iPhone 5. Each image is annotated with pixel-level binary segmentation masks. The cracks in this dataset primarily represent thin fissures on asphalt surfaces, presenting unique challenges distinct from those in the DeepCrack dataset. 

Crack500 Dataset: The Crack500 dataset contains 500 road crack images (2000×1500 pixels) collected at Temple University. Each image is split into 16 non-overlapping regions, totaling 3,368 sub-images, divided into 1,896 training, 1,124 test, and 384 validation images.

This combination of datasets allows for a thorough evaluation of the model's adaptability across different surface types and crack characteristics.

\subsection{Evaluation Metrics}
\textbf{Performance assessment.} To comprehensively evaluate the performance of the proposed model, we utilize three widely used and effective metrics for semantic segmentation: F1 Score, Intersection over Union (IoU), and Dice Coefficient.

The F1 Score represents the harmonic mean of Precision and Recall, combining the strengths of both metrics to provide a balanced measure of the model's accuracy and completeness in identifying cracks. It effectively quantifies the trade-off between false positives and false negatives. The F1 Score is defined as:

\[
F1 = 2 \times \frac{\text{Precision} \times \text{Recall}}{\text{Precision} + \text{Recall}}~,
\]

IoU is a critical metric for evaluating the overlap between the predicted crack results and the ground truth. It measures how well the predicted region aligns with the true region, providing a robust assessment of segmentation accuracy.
The formula is as follows:
\begin{equation}
IoU=  \frac {TP}{TP+FN+FP} ~,
\end{equation}
where $TP$ indicates true positive, $FN$ represents false negative, and $FP$ is false positive.

The Dice Coefficient emphasizes the ratio of twice the overlapping area to the total area of both the predicted and ground truth regions. This metric is particularly well-suited for evaluating the performance of segmentation tasks involving small objects, as it is more sensitive to slight variations in overlap. 
The formula is as follows:

\begin{equation}
Dice=  \frac {2\times TP}{2\times TP + FN + FP} ~.
\end{equation}

\textbf{Efficiency assessment.} To assess the efficiency of the proposed model, we employ three critical metrics: model parameters, inference time, and GFLOPs. 

Model Parameters: Measured in millions (M), this metric quantifies the total number of trainable parameters in the model, reflecting its structural complexity. Fewer parameters typically translate to reduced memory usage and storage requirements, which are essential for deployment on resource-limited devices.

Inference Time: Expressed in milliseconds (ms), inference time indicates the average duration required to process a single input during prediction. Faster inference times are crucial for real-time applications, particularly in environments where rapid decision-making is necessary.

GFLOPs: This metric represents the number of billions of floating-point operations performed during inference, providing a measure of the computational complexity. Lower GFLOPs indicate greater computational efficiency, making a model more suitable for resource-constrained environments.

By evaluating these metrics collectively, we gain a comprehensive understanding of the model’s computational demands, runtime performance, and overall feasibility for deployment in practical, real-world scenarios.

\subsection{Experimental results}
To evaluate the performance of the proposed model, we conducted a comparative analysis against twelve state-of-the-art methods commonly used in crack segmentation. These include three classical encoder-decoder architectures, FCN \cite{long2015fully} and U-Net \cite{ronneberger2015u}, along with SegNet \cite{badrinarayanan2017segnet}. Additionally, we considered PSP-Net \cite{zhao2017pyramid}, DeepLabv3+\cite{chen2018encoder}
and EMCAD\cite{rahman2024emcad}. The comparison also included the transformer-based model Segformer \cite{xie2021segformer}, HRViT\cite{gu2022multi}, CMTFNet\cite{wu2023cmtfnet}, and four crack-specific methods: DeepCrack \cite{zou2018deepcrack}, which leverages hierarchical feature extraction, DeepCrackAT \cite{lin2023deepcrackat}, integrating attention mechanisms for boundary refinement, and two recent methods, CT-CrackSeg \cite{tao2023convolutional} and CrackMamba\cite{ZUO2024105845}. 

For a comprehensive evaluation, twelve of these models were trained on the DeepCrack training set and subsequently evaluated on both the DeepCrack test set, the entire CFD dataset and Crack500 dataset. This dual-dataset evaluation ensures a robust comparison of performance across diverse crack segmentation scenarios.

\textbf{Comparison on the DeepCrack test set.} 
As shown in Table \ref{tabel1}, our method achieves significant improvements over existing approaches in crack segmentation. Among crack-specific methods, CrackMamba\cite{ZUO2024105845} ranks second; however, our approach surpasses it with notable gains of 1.47\% in F1 score, 2.56\% in IoU, and 2.0\% in Dice coefficient, demonstrating superior performance in crack detection and segmentation. Additionally, our method outperforms all general-purpose segmentation models in the comparison. Specifically, compared to SegNet \cite{badrinarayanan2017segnet}, our approach achieves improvements of 2.29\% in F1 score, 3.33\% in IoU, and 2.79\% in Dice, further highlighting its advantage over non-specialized models.

These results underscore the effectiveness of a tailored approach designed to address the unique challenges of crack detection. The substantial performance gains across key metrics, including F1 score, IoU, and Dice, emphasize the robustness and reliability of our method for real-world applications, where accurate segmentation is crucial. The consistent outperformance of both crack-specific and general-purpose models further validates the importance of domain-specific approaches in tackling specialized segmentation tasks.

\begin{table*}
\renewcommand{\arraystretch}{1.5}
\setlength{\tabcolsep}{5pt}
\normalsize

\caption{Comparison on the DeepCrack test dataset and zero-shot on the CFD dataset (\%)}
\label{tabel1}
\centering
\begin{tabular}{c|ccc|ccc|c|c|c}
\hline
\multirow{2}{*}{Method} & \multicolumn{3}{c|}{DeepCrack Test set} & \multicolumn{3}{c|}{CFD  Dataset} & \multirow{2}{*}{\shortstack{Model \\Params(M)}} & \multirow{2}{*}{\shortstack{Inference\\ Time (ms)}} & \multirow{2}{*}{\shortstack{GFLOPs}} \\ \cline{2-7}
                        & F1      & IoU      & Dice    & F1     & IoU    & Dice    &                              &                                     &                \\ \hline
FCN\cite{long2015fully}                     & 78.77   & 67.42    & 78.12   & 34.73  & 23.81  & 34.78   & 18.64                        & \textbf{4.59}                                & 101.99         \\
PSP-Net\cite{zhao2017pyramid}                 & 77.76   & 64.24    & 76.29   & 45.94  & 31.97  & 44.89   & 46.71                        & 9.91                                & 184.73         \\ 
DeepLabv3+\cite{chen2018encoder}              & 75.98   & 63.58    & 74.89   & 35.80  & 25.40  & 36.18   & 59.34                        & 8.81                                & 88.95          \\
SegNet\cite{badrinarayanan2017segnet}                  & 80.59   & 68.00    & 79.46   & 43.84  & 30.98  & 43.50   & 29.44                        & 8.37                                & 160.52         \\
Segformer\cite{xie2021segformer}               & 78.71   & 65.44    & 77.51   & 40.08  & 27.59  & 39.67   & 63.99                        & 50.68                               & \textbf{85.41}          \\
HRViT\cite{gu2022multi}                  & 79.58   & 66.80    & 78.70 & 33.35   & 22.65    & 33.32  & 26.02                      & 96.31                                & 28.20  \\
CMTFNet\cite{wu2023cmtfnet} & 76.81   & 62.33    & 74.48                 & 23.41   & 16.34    & 23.97 & 30.07                      & 10.73                                & 33.05     \\
EMCAD\cite{rahman2024emcad} & 79.18   & 66.60    & 78.09                & 46.96   & 31.73    & 46.03 & 83.35                     & 30.10                                 & 61.47     \\
\hline
\multicolumn{4}{l}{Crack-Specific Methods} \\
\hline
DeepCrack\cite{zou2018deepcrack}               & 78.95   & 66.49    & 77.85   & 36.97  & 25.25  & 36.83   & 30.91                        & 25.09                               & 547.22         \\
DeepCrackAT\cite{lin2023deepcrackat}             & 79.50   & 63.72    & 75.96   & 38.69  & 23.51  & 34.66   & \textbf{13.33}                        & 98.64                               & 374.88         \\
CT-CrackSeg\cite{tao2023convolutional}               & 79.22   & 67.25    & 78.30    & 30.99   & 21.64    & 30.89 & 22.88                        & 116.97                               & 163.33 \\
% DTrC-Net\cite{xiang2023crack}               & 81.39   & 69.38    & 80.69 & 38.17   & 26.15    & 38.06  & 63.45      & 23.61                               & 123.20 \\
CrackMamba\cite{ZUO2024105845}               & 81.41   & 68.77   & 80.25 &   49.98    & 34.40  & 49.25      & 75.95 & 84.73                                & 192.27 \\
\hline
\textbf{Ours}           & \textbf{82.88} & \textbf{71.33} & \textbf{82.25} & \textbf{54.48} & \textbf{39.15} & \textbf{53.83} & 25.63     & \textbf{22.09}  & 196.18 \\ \hline
\end{tabular}
\end{table*}

\textbf{Zero-shot evaluation on the CFD dataset.} 
As shown in Table \ref{tabel1}, under zero-shot cross-domain evaluation on the CFD dataset, our method outperforms all competing approaches across all three performance metrics: F1 score, IoU, and Dice coefficient. Notably, it surpasses the second-best method, CrackMamba\cite{ZUO2024105845}, by substantial margins, achieving improvements of 4.5\% in F1 score, 4.75\% in IoU, and 4.58\% in Dice coefficient. Additionally, our approach demonstrates superior performance over other state-of-the-art models, including EMCAD\cite{rahman2024emcad} and DeepCrackAT \cite{lin2023deepcrackat}, with significant gains across all metrics. For instance, compared to EMCAD\cite{rahman2024emcad}, our method achieves improvements of 7.52\% in F1 score, 7.42\% in IoU, and 7.8\% in Dice coefficient.

\textbf{Zero-shot evaluation on the Crack500 dataset.} To further assess the generalization capability of our model, we conducted a zero-shot quantitative evaluation on the Crack500 dataset. As shown in Table \ref{tabel2}, our method once again achieved the best performance compared to other approaches. Specifically, on the validation set, our method outperformed the second-best performing method, CMTFNet, by 8.09\%, 10.99\%, and 10.64\% in terms of F1 score, IoU, and Dice coefficient, respectively. On the test set, our method surpassed the second-best method, CrackMamba, by 2.45\%, 1.06\%, and 1.84\% in the same metrics.

These results underscore the strong generalization capability of our model when applied to cross-domain datasets, highlighting its effectiveness in addressing diverse and challenging real-world conditions. The substantial performance gains reinforce the robustness and adaptability of our approach, establishing it as a powerful solution for crack segmentation in a wide range of practical scenarios.

\begin{table}
\renewcommand{\arraystretch}{1.5}
\setlength{\tabcolsep}{2.5pt}
\normalsize

\caption{Zero-shot on the Crack500 dataset (\%)}
\label{tabel2}
\centering
\begin{tabular}{c|ccc|ccc}
\hline
\multirow{2}{*}{Method} & \multicolumn{3}{c|}{val set} & \multicolumn{3}{c}{test set} \\ \cline{2-7}
                        & F1      & IoU      & Dice    & F1     & IoU    & Dice    \\ \hline
FCN\cite{long2015fully}                     & 39.49   & 27.44   & 38.98   & 31.35  & 20.26  & 30.91   \\
PSP-Net\cite{zhao2017pyramid}                 & 34.66   & 24.76    & 34.58   & 29.00  & 19.65  & 28.98   \\ 
DeepLabv3+\cite{chen2018encoder}              & 32.83   & 23.55    & 32.83   & 14.88  & 10.34  & 15.62  \\
SegNet\cite{badrinarayanan2017segnet}                  & 42.32   & 30.18    & 41.80   & 36.27  & 24.45  & 35.59   \\
Segformer\cite{xie2021segformer}               & 46.29   & 33.10    & 45.74   & 41.38  & 28.10  & 40.75   \\
HRViT\cite{gu2022multi}                  & 40.35   & 28.13    & 40.13 & 33.32   & 21.86    & 32.89  \\
CMTFNet\cite{wu2023cmtfnet} & 48.76   & 31.91    & 45.39                 & 41.21   & 24.30    & 36.82 \\
EMCAD\cite{rahman2024emcad} & 44.37   & 30.94     & 43.50             & 40.88   & 27.06    & 39.59 \\
\hline
\multicolumn{4}{l}{Crack-Specific Methods} \\
\hline
DeepCrack\cite{zou2018deepcrack}               & 39.10   & 26.22    & 38.23   & 30.94  & 19.30  & 30.04   \\
DeepCrackAT\cite{lin2023deepcrackat}             & 43.80   & 28.46    & 41.27   & 37.55  & 22.82  & 34.97   \\
CT-CrackSeg\cite{tao2023convolutional}               & 44.10   & 31.93    & 43.80   & 42.03   & 29.11    & 41.56 \\

CrackMamba\cite{ZUO2024105845}               & 47.43   & 34.44 & 47.28 &   42.56    & 29.34  & 42.40      \\
\hline
\textbf{Ours}           & \textbf{56.85} & \textbf{42.30} & \textbf{56.03} & \textbf{45.01} & \textbf{30.40} & \textbf{44.24} \\ \hline
\end{tabular}
\end{table}

\textbf{Efficiency comparison.} 
As shown in Table \ref{tabel1}, we compare the efficiency and complexity of various models using three key metrics: model parameters, inference speed, and GFLOPs, which serve as an auxiliary indicator of computational complexity. The results clearly demonstrate that, despite incorporating a general feature extractor module, our method introduces only minimal additional computational cost compared to advanced models in the crack segmentation domain. Remarkably, our approach surpasses state-of-the-art crack segmentation models in terms of efficiency.

Specifically, when compared to the previous leading model, CrackMamba\cite{ZUO2024105845}, our method achieves significantly fewer parameters, faster inference speeds, and superior performance metrics, resulting in a solution that is not only smaller and faster but also more accurate. In terms of inference speed (measured in milliseconds), our model outperforms all competing crack segmentation methods, achieving a processing time that is 62.64 ms faster than CrackMamba, and 76.55 ms faster than DeepCrackAT \cite{lin2023deepcrackat}. This substantial improvement in processing speed is critical for real-world deployment, where rapid crack detection is essential for timely analysis and decision-making.

Moreover, our method demonstrates a significant advantage in terms of model parameters compared to the second-best CrackMamba \cite{ZUO2024105845}, as indicated by the Model Params metric. Specifically, our model achieves this with 25.63M parameters, which is significantly smaller than the 75.95M parameters of CrackMamba. These results underscore the practicality of our approach for real-world applications, where model size and inference speed are critical considerations.

\begin{figure*}
\centering
\subfigure[Images]{
\begin{minipage}{0.12\linewidth}
    \centering
    \includegraphics[width=2.0cm,height=2.0cm]{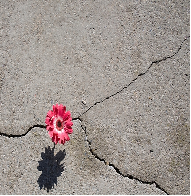}\\
    \vspace{0.1cm}
    \includegraphics[width=2.0cm,height=2.0cm]{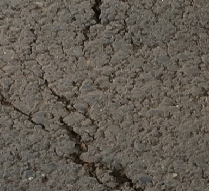}\\
    \vspace{0.1cm}
    \includegraphics[width=2.0cm,height=2.0cm]{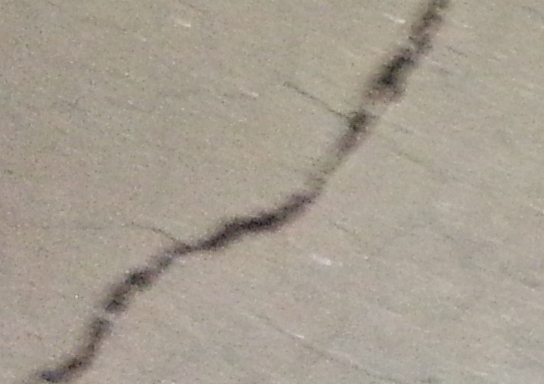}\\
    \vspace{0.1cm}
    \includegraphics[width=2.0cm,height=2.0cm]{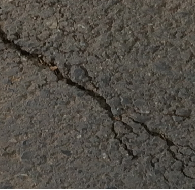}\\
    \vspace{0.1cm}
    \includegraphics[width=2.0cm,height=2.0cm]{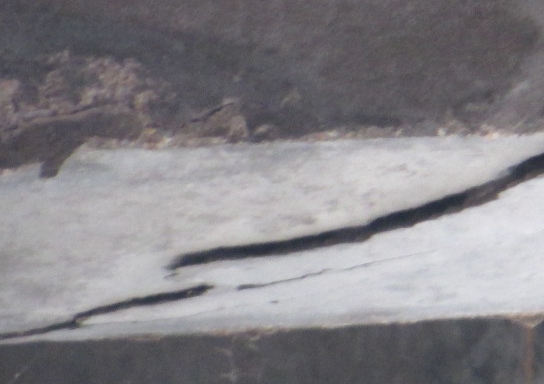}\\
    \vspace{0.1cm}
\end{minipage}\hspace{-3mm}
}%
\subfigure[Ground Truth]{
\begin{minipage}{0.12\linewidth}
    \centering
    \includegraphics[width=2.0cm,height=2.0cm]{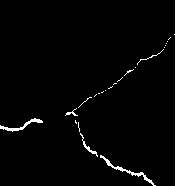}\\
    \vspace{0.1cm}
    \includegraphics[width=2.0cm,height=2.0cm]{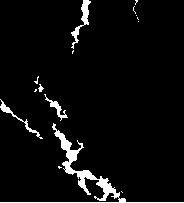}\\
    \vspace{0.1cm}
    \includegraphics[width=2.0cm,height=2.0cm]{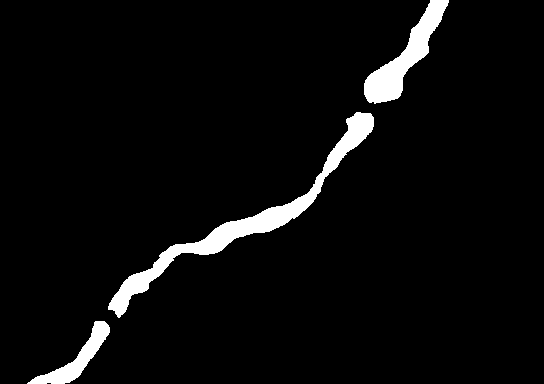}\\
    \vspace{0.1cm}
    \includegraphics[width=2.0cm,height=2.0cm]{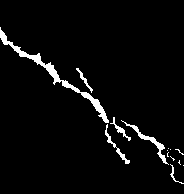}\\
    \vspace{0.1cm}
    \includegraphics[width=2.0cm,height=2.0cm]{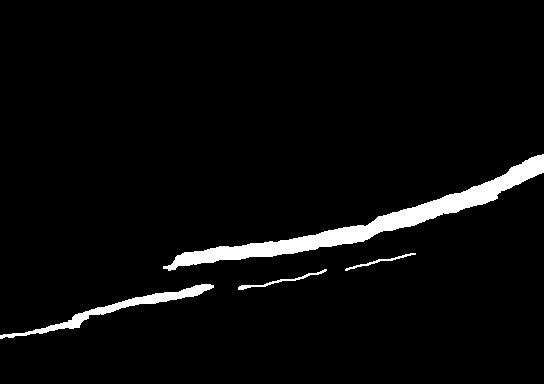}\\
    \vspace{0.1cm}
\end{minipage}\hspace{-3mm}
}%
\subfigure[SegNet]{
\begin{minipage}{0.12\linewidth}
    \centering
    \includegraphics[width=2.0cm,height=2.0cm]{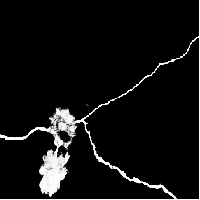}\\
    \vspace{0.1cm}
    \includegraphics[width=2.0cm,height=2.0cm]{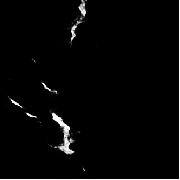}\\
    \vspace{0.1cm}
    \includegraphics[width=2.0cm,height=2.0cm]{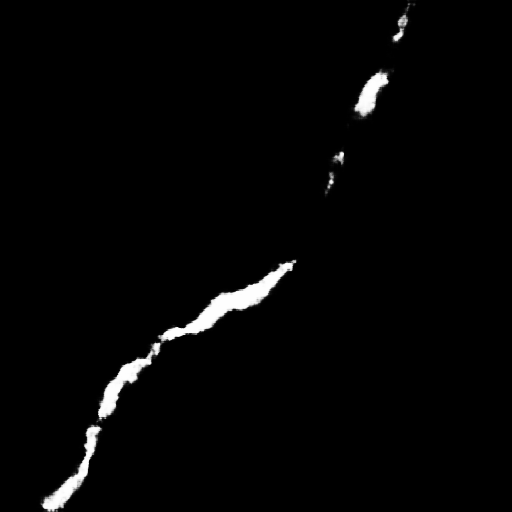}\\
    \vspace{0.1cm}
    \includegraphics[width=2.0cm,height=2.0cm]{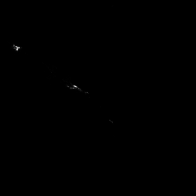}\\
    \vspace{0.1cm}
    \includegraphics[width=2.0cm,height=2.0cm]{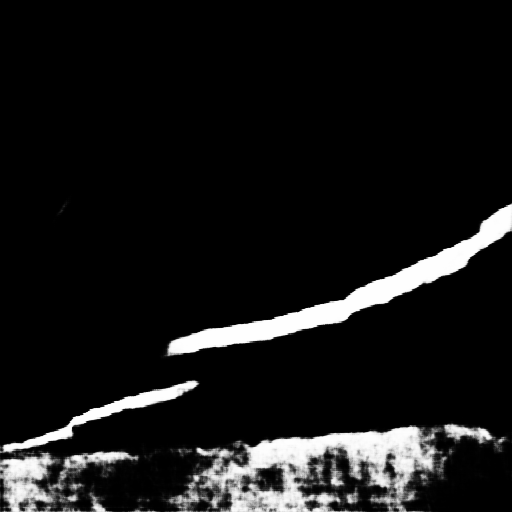}\\
    \vspace{0.1cm}
\end{minipage}\hspace{-2mm}%
}%
\subfigure[HRViT]{
\begin{minipage}{0.12\linewidth}
    \centering
    \includegraphics[width=2.0cm,height=2.0cm]{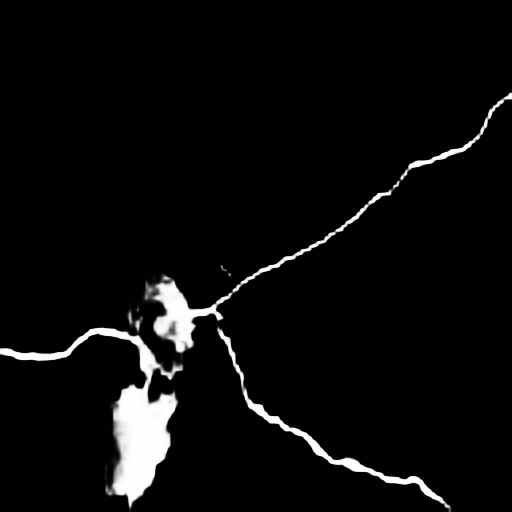}\\
    \vspace{0.1cm}
    \includegraphics[width=2.0cm,height=2.0cm]{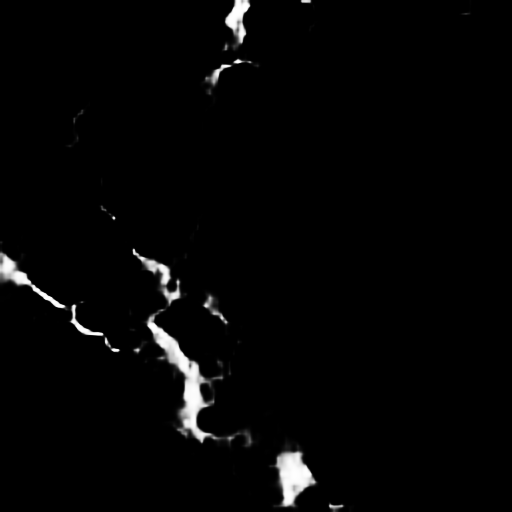}\\
    \vspace{0.1cm}
    \includegraphics[width=2.0cm,height=2.0cm]{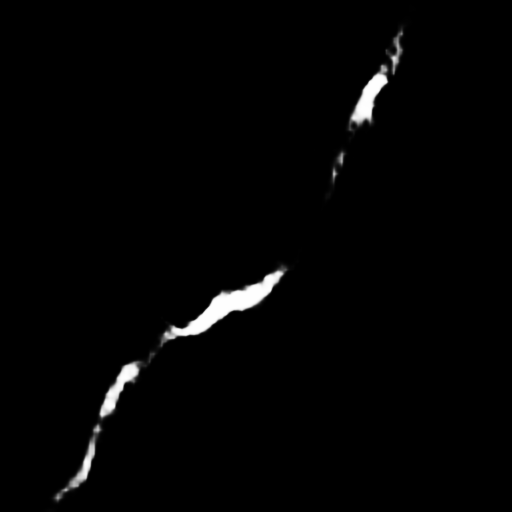}\\
    \vspace{0.1cm}
    \includegraphics[width=2.0cm,height=2.0cm]{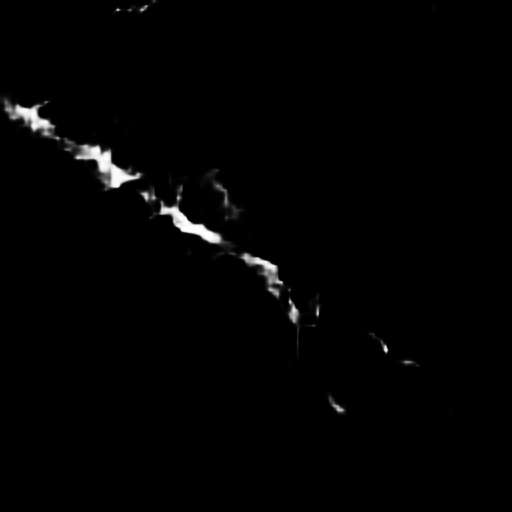}\\
    \vspace{0.1cm}
    \includegraphics[width=2.0cm,height=2.0cm]{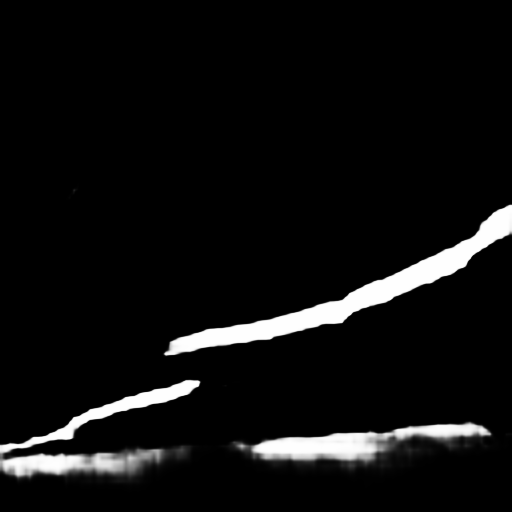}\\
    \vspace{0.1cm}
\end{minipage}\hspace{-2mm}%
}%
\subfigure[DeepCrackAT]{
\begin{minipage}{0.12\linewidth}
    \centering
    \includegraphics[width=2.0cm,height=2.0cm]{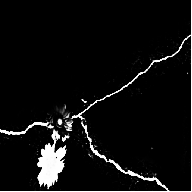}\\
    \vspace{0.1cm}
    \includegraphics[width=2.0cm,height=2.0cm]{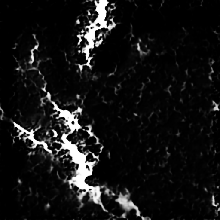}\\
    \vspace{0.1cm}
    \includegraphics[width=2.0cm,height=2.0cm]{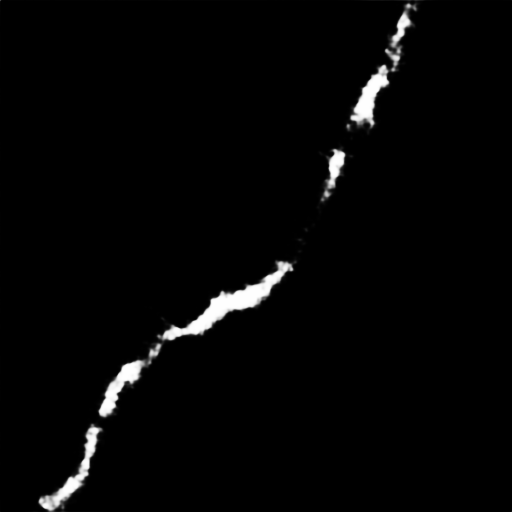}\\
    \vspace{0.1cm}
    \includegraphics[width=2.0cm,height=2.0cm]{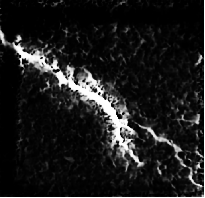}\\
    \vspace{0.1cm}
    \includegraphics[width=2.0cm,height=2.0cm]{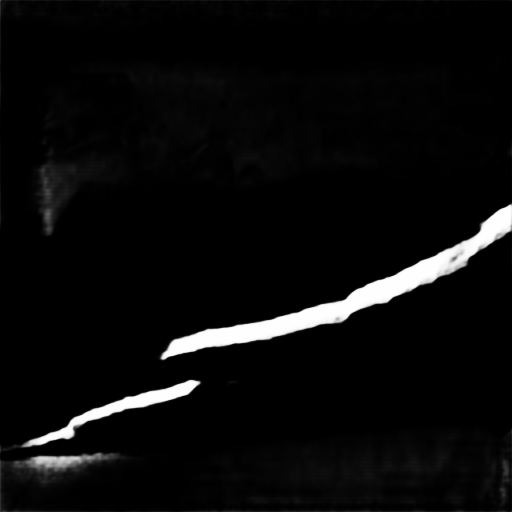}\\
    \vspace{0.1cm}
\end{minipage}\hspace{-2mm}%
}%
\subfigure[CT-CrackSeg]{
\begin{minipage}{0.12\linewidth}
    \centering
    \includegraphics[width=2.0cm,height=2.0cm]{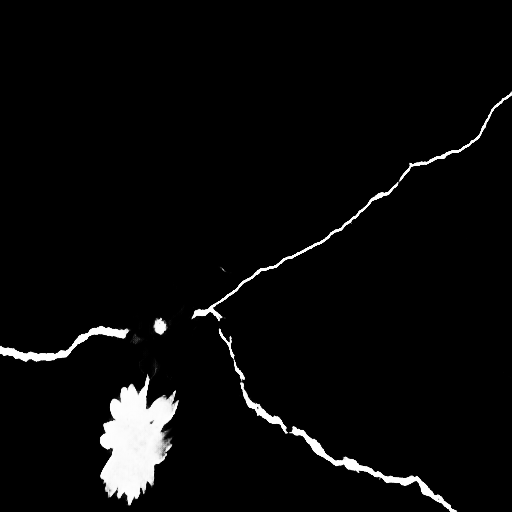}\\
    \vspace{0.1cm}
    \includegraphics[width=2.0cm,height=2.0cm]{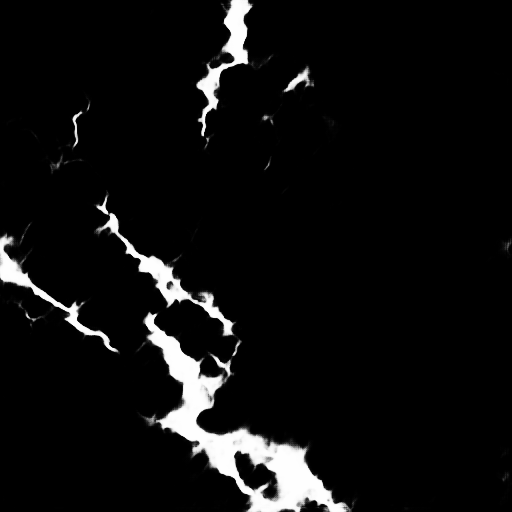}\\
    \vspace{0.1cm}
    \includegraphics[width=2.0cm,height=2.0cm]{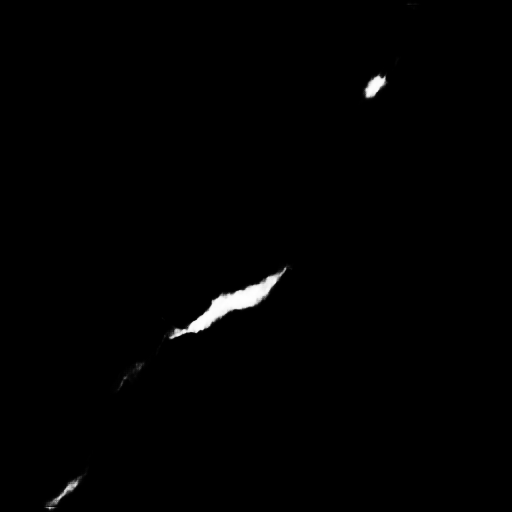}\\
    \vspace{0.1cm}
    \includegraphics[width=2.0cm,height=2.0cm]{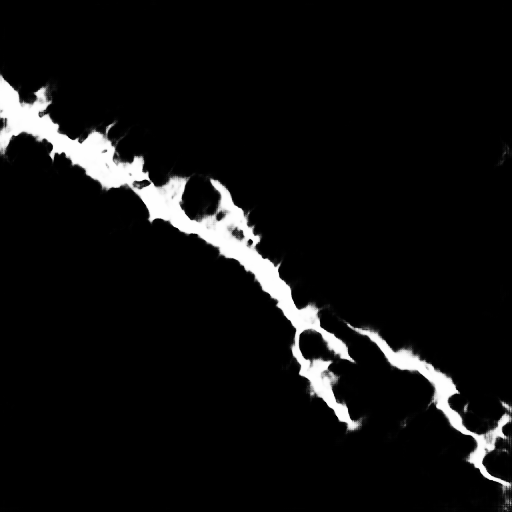}\\
    \vspace{0.1cm}
    \includegraphics[width=2.0cm,height=2.0cm]{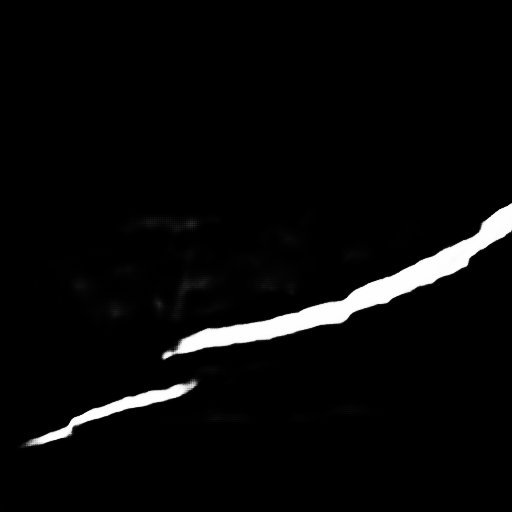}\\
    \vspace{0.1cm}
\end{minipage}\hspace{-2mm}%
}%
\subfigure[CrackMamba]{
\begin{minipage}{0.12\linewidth}
    \centering
    \includegraphics[width=2.0cm,height=2.0cm]{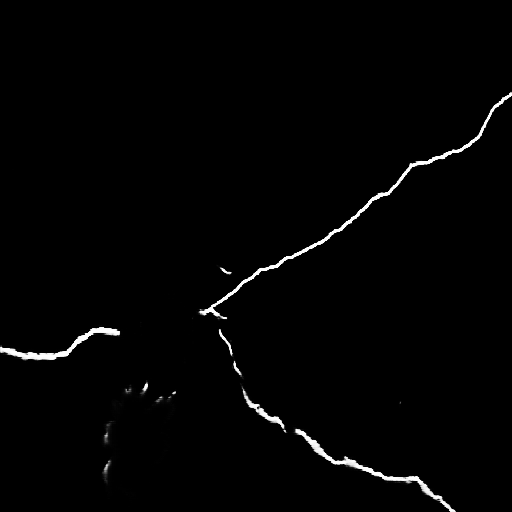}\\
    \vspace{0.1cm}
    \includegraphics[width=2.0cm,height=2.0cm]{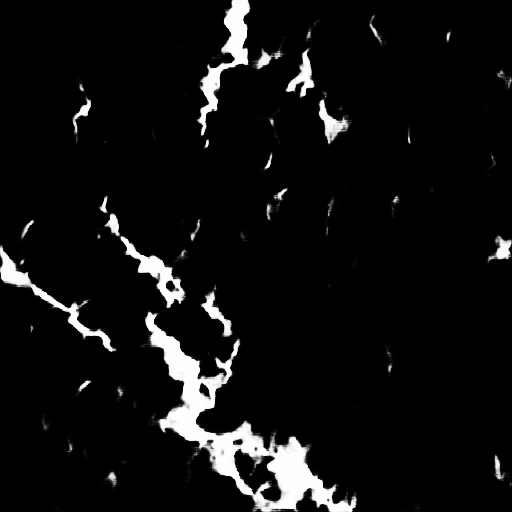}\\
    \vspace{0.1cm}
    \includegraphics[width=2.0cm,height=2.0cm]{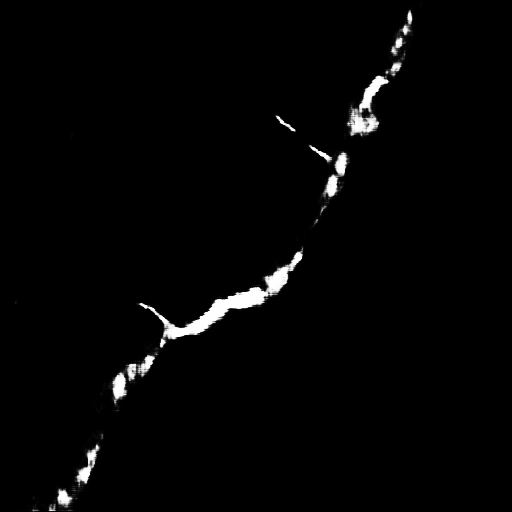}\\
    \vspace{0.1cm}
    \includegraphics[width=2.0cm,height=2.0cm]{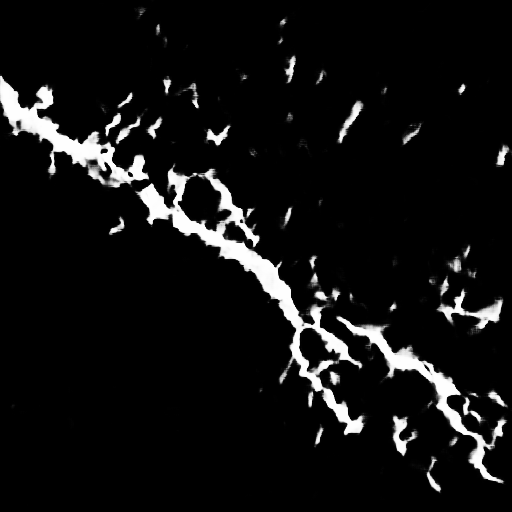}\\
    \vspace{0.1cm}
    \includegraphics[width=2.0cm,height=2.0cm]{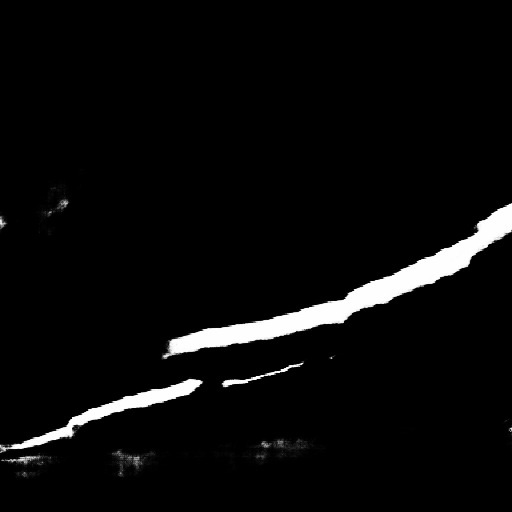}\\
    \vspace{0.1cm}
\end{minipage}\hspace{-2mm}%
}%
\subfigure[Ours]{
\begin{minipage}{0.12\linewidth}
    \centering
    \includegraphics[width=2.0cm,height=2.0cm]{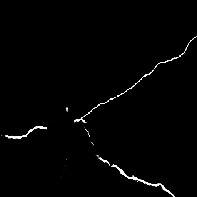}\\
    \vspace{0.1cm}
    \includegraphics[width=2.0cm,height=2.0cm]{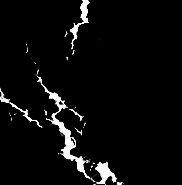}\\
    \vspace{0.1cm}
    \includegraphics[width=2.0cm,height=2.0cm]{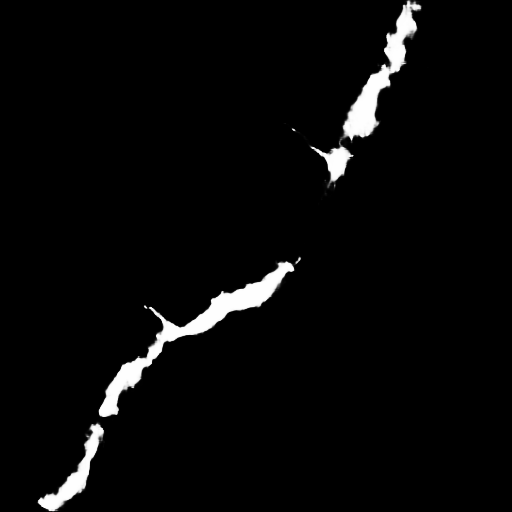}\\
    \vspace{0.1cm}
    \includegraphics[width=2.0cm,height=2.0cm]{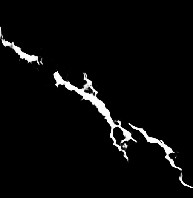}\\
    \vspace{0.1cm}
    \includegraphics[width=2.0cm,height=2.0cm]{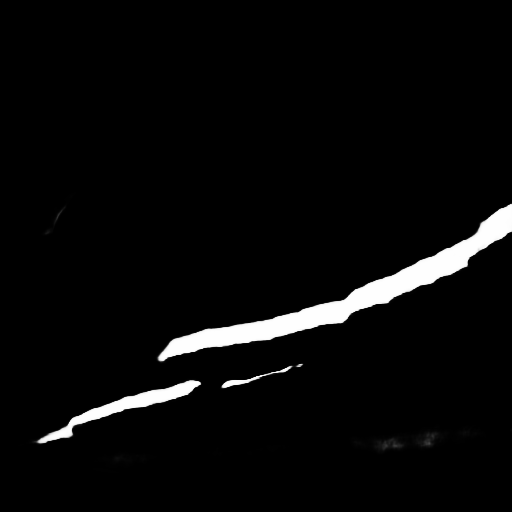}\\
    \vspace{0.1cm}
\end{minipage}
}
\caption{Visualization of samples from the DeepCrack test set.}
\label{fig5}
\end{figure*}

% We also conducted a comparison with SAM \cite{kirillov2023segment}, and the results underscore the efficiency of our proposed method. SAM has over 25 times more parameters, operates at 20 times slower inference speeds, and exhibits nearly 14 times higher computational complexity compared to our model. This substantial computational overhead poses significant challenges for deployment, particularly in mobile crack detection systems with limited resources. 

In contrast, our method integrates general segmentation knowledge while maintaining drastically lower inference costs. This efficiency makes our approach far more practical and suitable for deployment in resource-constrained environments, highlighting its advantages for real-world applications.

\subsection{Qualitative Evaluation}
\begin{figure*}
\centering
\subfigure[Images]{
\begin{minipage}{0.12\linewidth}
    \centering
    \includegraphics[width=2.0cm,height=2.0cm]{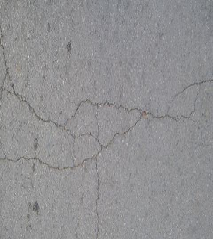}\\
    \vspace{0.1cm}
    \includegraphics[width=2.0cm,height=2.0cm]{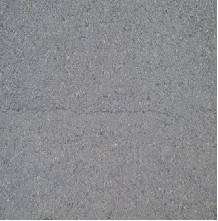}\\
    \vspace{0.1cm}
    \includegraphics[width=2.0cm,height=2.0cm]{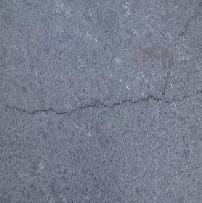}\\
    \vspace{0.1cm}
    \includegraphics[width=2.0cm,height=2.0cm]{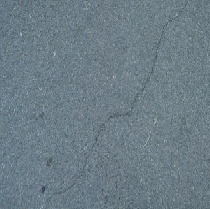}\\
    \vspace{0.1cm}
\end{minipage}\hspace{-3mm}
}%
\subfigure[Ground Truth]{
\begin{minipage}{0.12\linewidth}
    \centering
    \includegraphics[width=2.0cm,height=2.0cm]{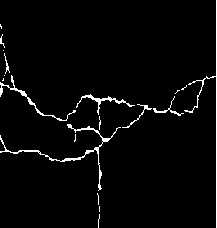}\\
    \vspace{0.1cm}
    \includegraphics[width=2.0cm,height=2.0cm]{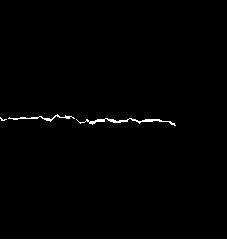}\\
    \vspace{0.1cm}
    \includegraphics[width=2.0cm,height=2.0cm]{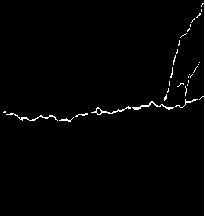}\\
    \vspace{0.1cm}
    \includegraphics[width=2.0cm,height=2.0cm]{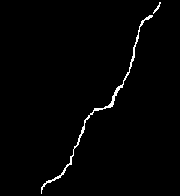}\\
    \vspace{0.1cm}
\end{minipage}\hspace{-3mm}
}%
\subfigure[PSP-Net]{
\begin{minipage}{0.12\linewidth}
    \centering
    \includegraphics[width=2.0cm,height=2.0cm]{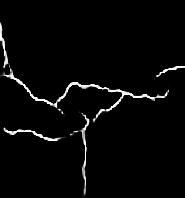}\\
    \vspace{0.1cm}
    \includegraphics[width=2.0cm,height=2.0cm]{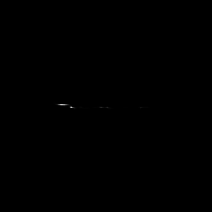}\\
    \vspace{0.1cm}
    \includegraphics[width=2.0cm,height=2.0cm]{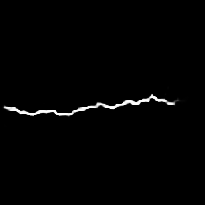}\\
    \vspace{0.1cm}
    \includegraphics[width=2.0cm,height=2.0cm]{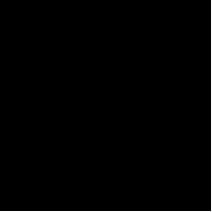}\\
    \vspace{0.1cm}
\end{minipage}\hspace{-3mm}
}%
\subfigure[SegNet]{
\begin{minipage}{0.12\linewidth}
    \centering
    \includegraphics[width=2.0cm,height=2.0cm]{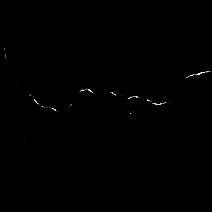}\\
    \vspace{0.1cm}
    \includegraphics[width=2.0cm,height=2.0cm]{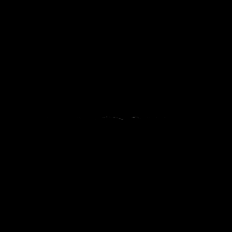}\\
    \vspace{0.1cm}
    \includegraphics[width=2.0cm,height=2.0cm]{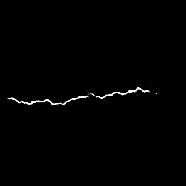}\\
    \vspace{0.1cm}
    \includegraphics[width=2.0cm,height=2.0cm]{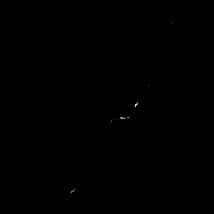}\\
    \vspace{0.1cm}
\end{minipage}\hspace{-3mm}
}%
\subfigure[Segformer]{
\begin{minipage}{0.12\linewidth}
    \centering
    \includegraphics[width=2.0cm,height=2.0cm]{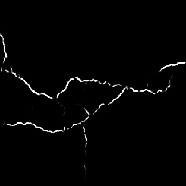}\\
    \vspace{0.1cm}
    \includegraphics[width=2.0cm,height=2.0cm]{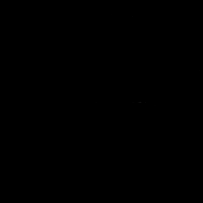}\\
    \vspace{0.1cm}
    \includegraphics[width=2.0cm,height=2.0cm]{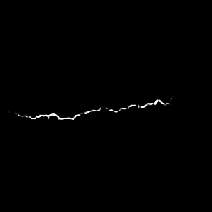}\\
    \vspace{0.1cm}
    \includegraphics[width=2.0cm,height=2.0cm]{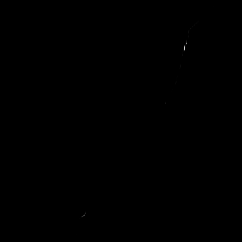}\\
    \vspace{0.1cm}
\end{minipage}\hspace{-3mm}
}%
\subfigure[DeepCrackAT]{
\begin{minipage}{0.12\linewidth}
    \centering
    \includegraphics[width=2.0cm,height=2.0cm]{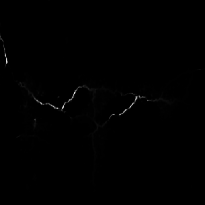}\\
    \vspace{0.1cm}
    \includegraphics[width=2.0cm,height=2.0cm]{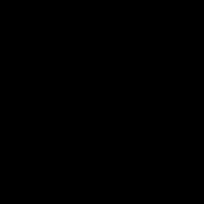}\\
    \vspace{0.1cm}
    \includegraphics[width=2.0cm,height=2.0cm]{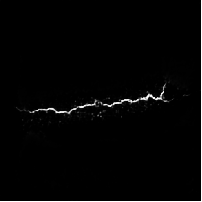}\\
    \vspace{0.1cm}
    \includegraphics[width=2.0cm,height=2.0cm]{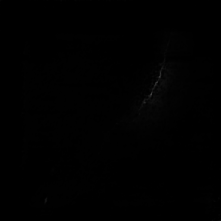}\\
    \vspace{0.1cm}
\end{minipage}\hspace{-3mm}
}%
\subfigure[CrackMamba]{
\begin{minipage}{0.12\linewidth}
    \centering
    \includegraphics[width=2.0cm,height=2.0cm]{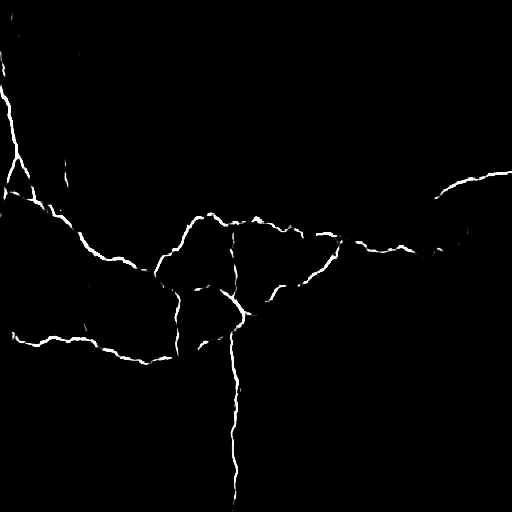}\\
    \vspace{0.1cm}
    \includegraphics[width=2.0cm,height=2.0cm]{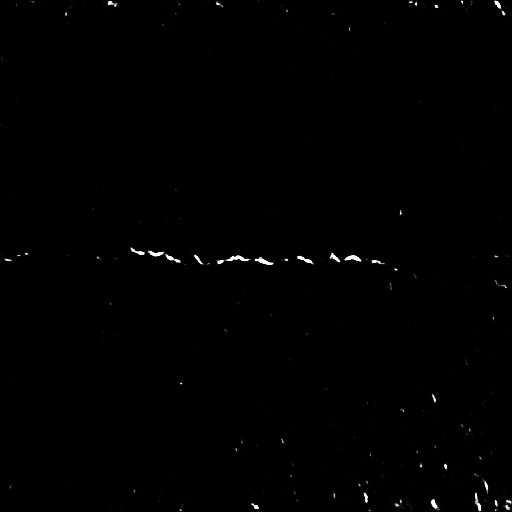}\\
    \vspace{0.1cm}
    \includegraphics[width=2.0cm,height=2.0cm]{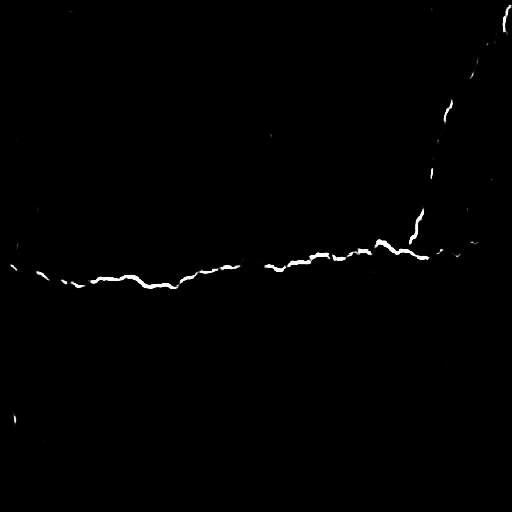}\\
    \vspace{0.1cm}
    \includegraphics[width=2.0cm,height=2.0cm]{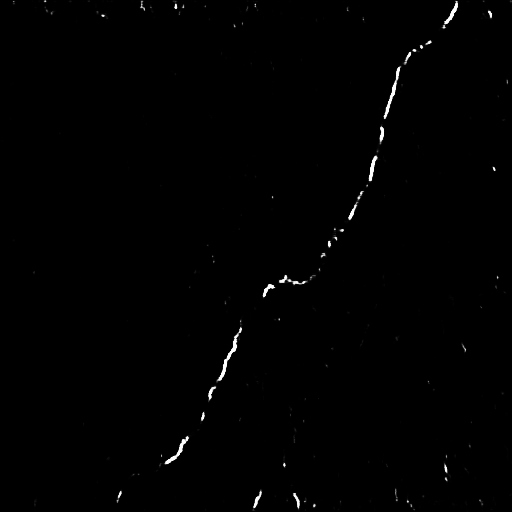}\\
    \vspace{0.1cm}
\end{minipage}\hspace{-3mm}
}%
\subfigure[Ours]{
\begin{minipage}{0.12\linewidth}
    \centering
    \includegraphics[width=2.0cm,height=2.0cm]{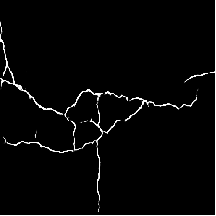}\\
    \vspace{0.1cm}
    \includegraphics[width=2.0cm,height=2.0cm]{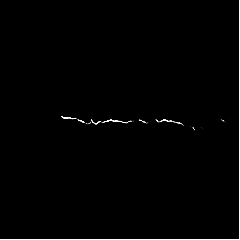}\\
    \vspace{0.1cm}
    \includegraphics[width=2.0cm,height=2.0cm]{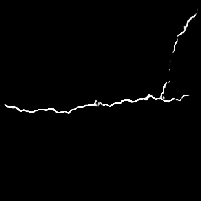}\\
    \vspace{0.1cm}
    \includegraphics[width=2.0cm,height=2.0cm]{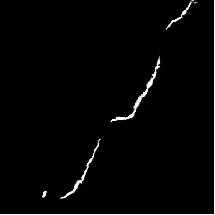}\\
    \vspace{0.1cm}
\end{minipage}\hspace{-3mm}
}%
\caption{Visualization of samples from the CFD dataset.}
\label{fig6}
\end{figure*}
\textbf{Visualization on the DeepCrack test set.} 
Fig. \ref{fig5} showcases the segmentation results on the DeepCrack test set, comparing the proposed method against five top-performing models. The results demonstrate the consistent superiority of our method, particularly under challenging conditions such as blurry inputs, complex backgrounds, and visually ambiguous artifacts.

In the first row, the input image features a crack alongside a flower and its shadow. 
While other methods misclassify the shadow as part of the crack—suggesting an over - reliance on local pixel contrasts rather than a comprehensive semantic understanding — FlexiCrackNet accurately excludes the shadow. This highlights its ability to capture high-level semantic features effectively. In the second row, a crack is presented on a highly textured and noisy surface. Other models output scattered or incomplete predictions, struggling to differentiate the crack from the background. In contrast, the proposed method suppresses irrelevant background features, delivering clean and precise segmentation. 

The third row presents a blurred image where the crack is faint and lacks distinct edges. Competing methods fail to maintain the continuity of the crack, resulting in fragmented predictions. However, the proposed method adapts well to the low-quality input, producing a coherent and continuous segmentation. In the fourth row, the crack is embedded in a surface with substantial noise and irregular textures. While other methods struggle with noise suppression and generate scattered outputs, the proposed method effectively isolates the crack with remarkable clarity and precision.

These results demonstrate the robustness of the proposed method in accurately detecting cracks across diverse and challenging scenarios. It consistently outperforms other approaches in suppressing noise, preserving crack continuity, and avoiding misclassification, achieving results that closely align with the ground truth.

\begin{figure*}
\centering
\subfigure[Images]{
\begin{minipage}{0.12\linewidth}
    \centering
    \includegraphics[width=2.0cm,height=2.0cm]{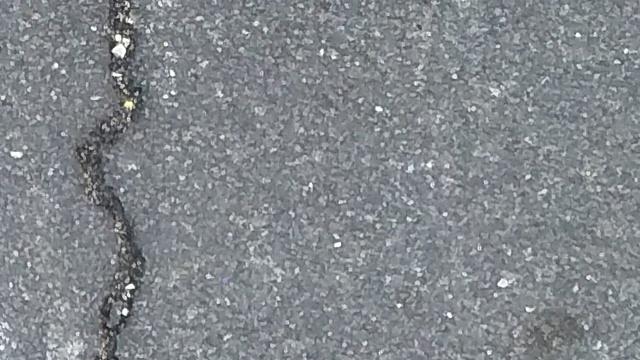}\\
    \vspace{0.1cm}
    \includegraphics[width=2.0cm,height=2.0cm]{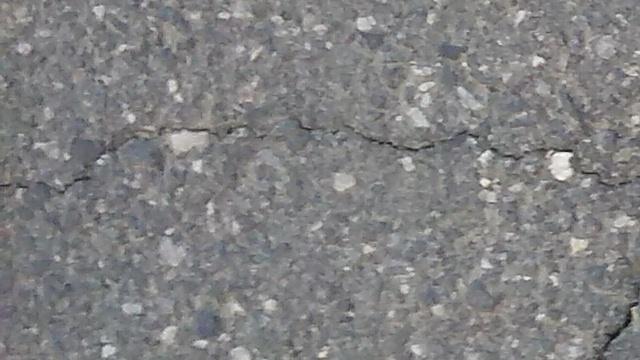}\\
    \vspace{0.1cm}
    \includegraphics[width=2.0cm,height=2.0cm]{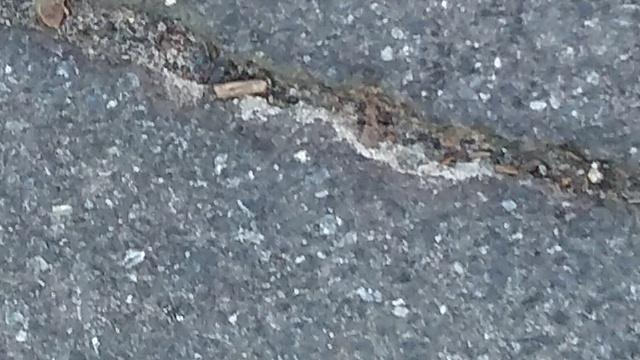}\\
    \vspace{0.1cm}
\end{minipage}\hspace{-3mm}
}%
\subfigure[Ground Truth]{
\begin{minipage}{0.12\linewidth}
    \centering
    \includegraphics[width=2.0cm,height=2.0cm]{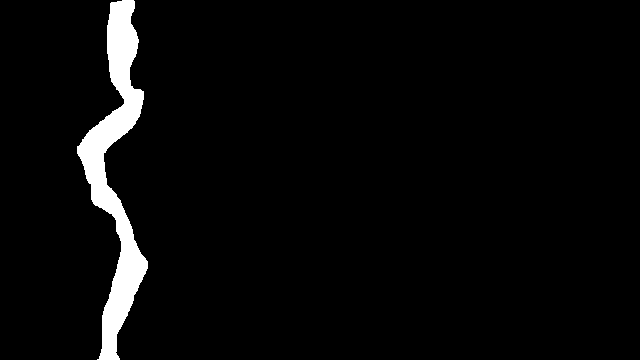}\\
    \vspace{0.1cm}
    \includegraphics[width=2.0cm,height=2.0cm]{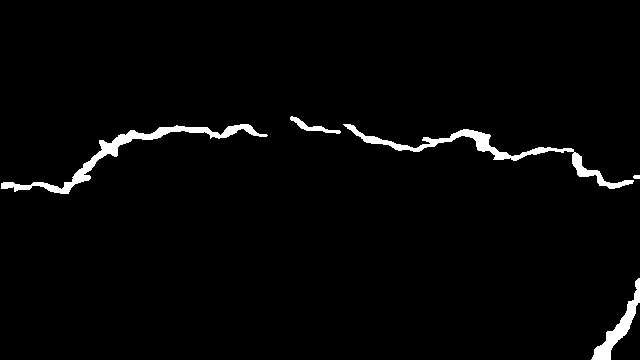}\\
    \vspace{0.1cm}
    \includegraphics[width=2.0cm,height=2.0cm]{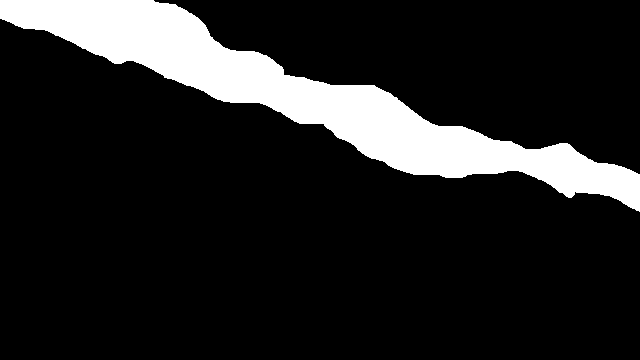}\\
    \vspace{0.1cm}
\end{minipage}\hspace{-3mm}
}%
\subfigure[EMCAD]{
\begin{minipage}{0.12\linewidth}
    \centering
    \includegraphics[width=2.0cm,height=2.0cm]{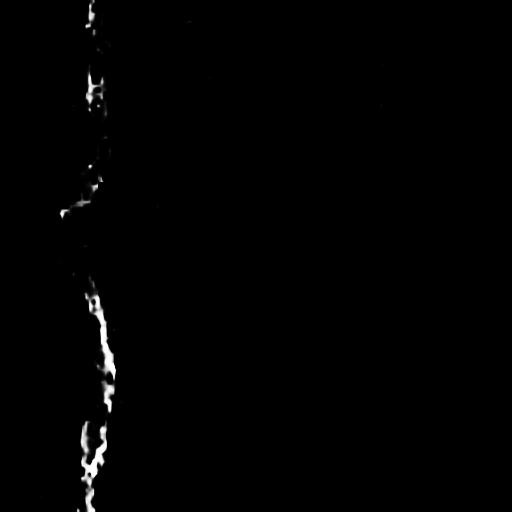}\\
    \vspace{0.1cm}
    \includegraphics[width=2.0cm,height=2.0cm]{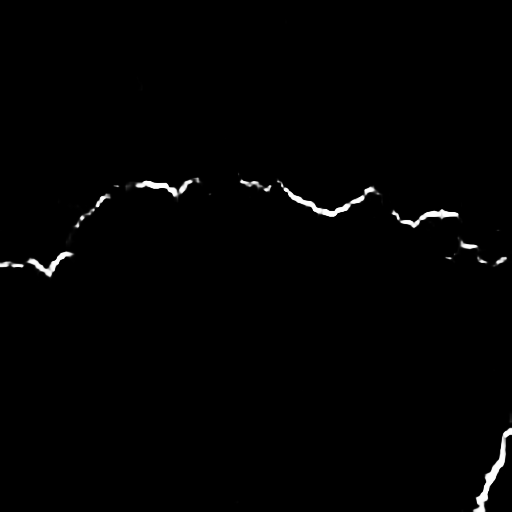}\\
    \vspace{0.1cm}
    \includegraphics[width=2.0cm,height=2.0cm]{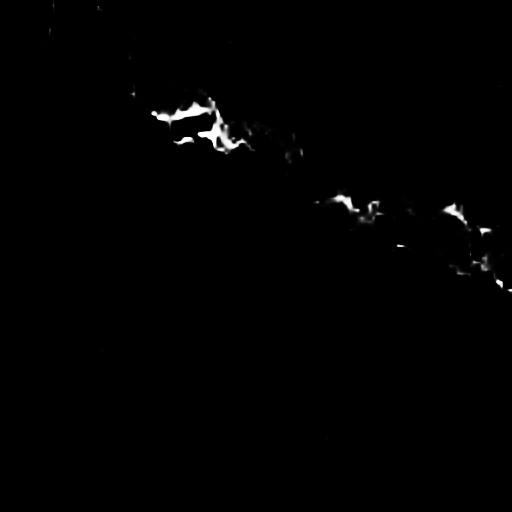}\\
    \vspace{0.1cm}
\end{minipage}\hspace{-3mm}
}%
\subfigure[CMTFNet]{
\begin{minipage}{0.12\linewidth}
    \centering
    \includegraphics[width=2.0cm,height=2.0cm]{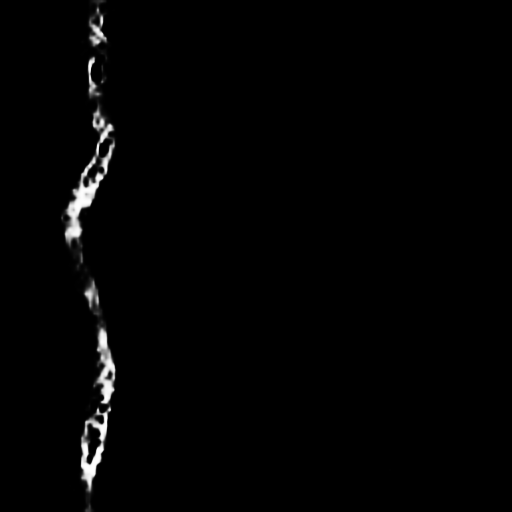}\\
    \vspace{0.1cm}
    \includegraphics[width=2.0cm,height=2.0cm]{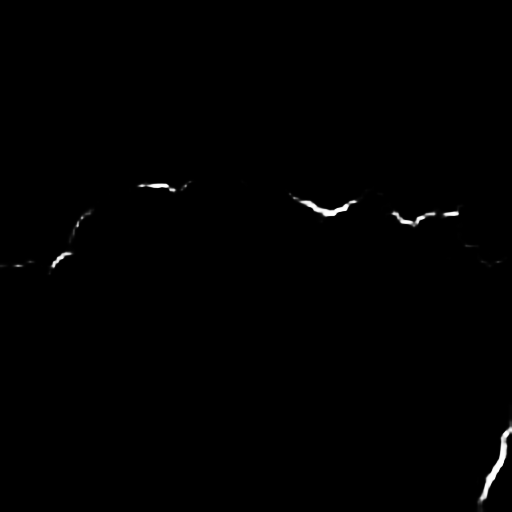}\\
    \vspace{0.1cm}
    \includegraphics[width=2.0cm,height=2.0cm]{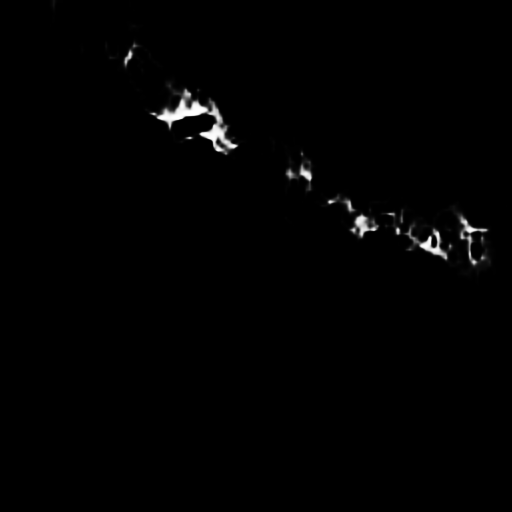}\\
    \vspace{0.1cm}
\end{minipage}\hspace{-3mm}
}%
\subfigure[EMCAD]{
\begin{minipage}{0.12\linewidth}
    \centering
    \includegraphics[width=2.0cm,height=2.0cm]{figure/crack500/1/EMCADNet.png}\\
    \vspace{0.1cm}
    \includegraphics[width=2.0cm,height=2.0cm]{figure/crack500/2/EMCADNet.png}\\
    \vspace{0.1cm}
    \includegraphics[width=2.0cm,height=2.0cm]{figure/crack500/3/EMCADNet.png}\\
    \vspace{0.1cm}
\end{minipage}\hspace{-3mm}
}%
\subfigure[CT-CrackSeg]{
\begin{minipage}{0.12\linewidth}
    \centering
    \includegraphics[width=2.0cm,height=2.0cm]{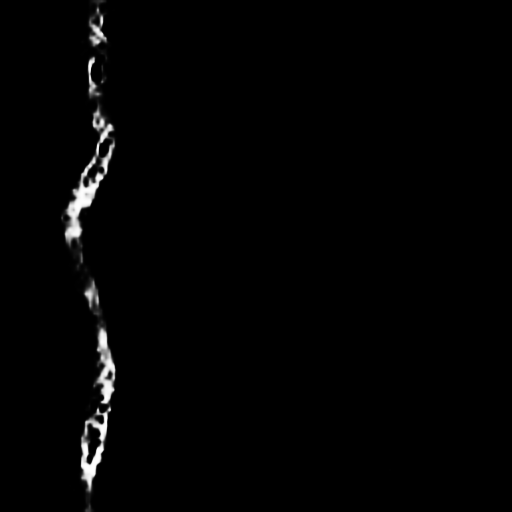}\\
    \vspace{0.1cm}
    \includegraphics[width=2.0cm,height=2.0cm]{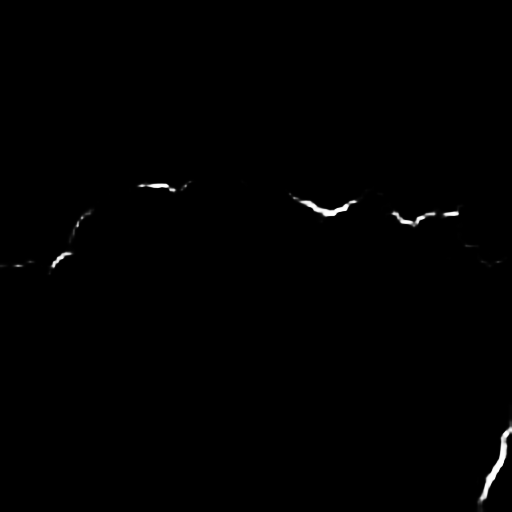}\\
    \vspace{0.1cm}
    \includegraphics[width=2.0cm,height=2.0cm]{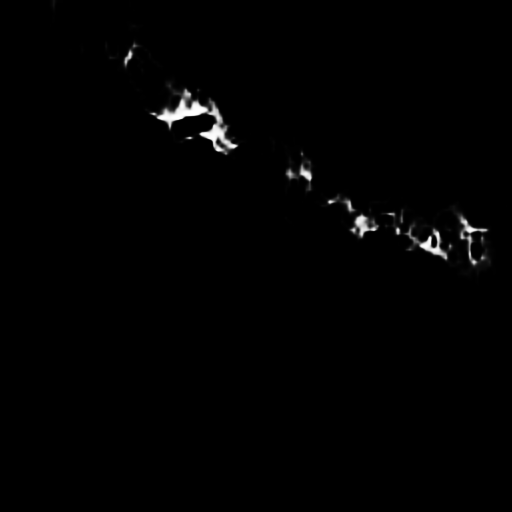}\\
    \vspace{0.1cm}
\end{minipage}\hspace{-3mm}
}%
\subfigure[CrackMamba]{
\begin{minipage}{0.12\linewidth}
    \centering
    \includegraphics[width=2.0cm,height=2.0cm]{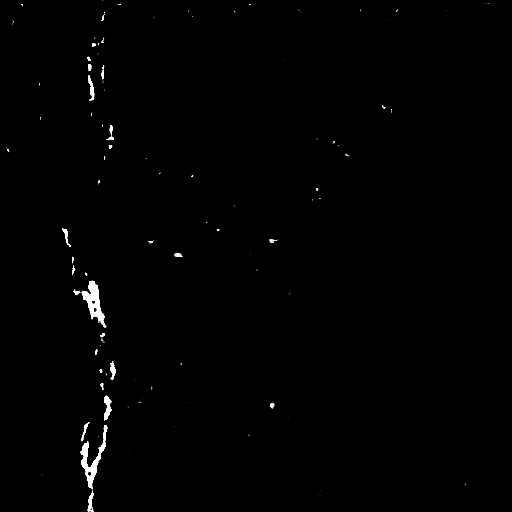}\\
    \vspace{0.1cm}
    \includegraphics[width=2.0cm,height=2.0cm]{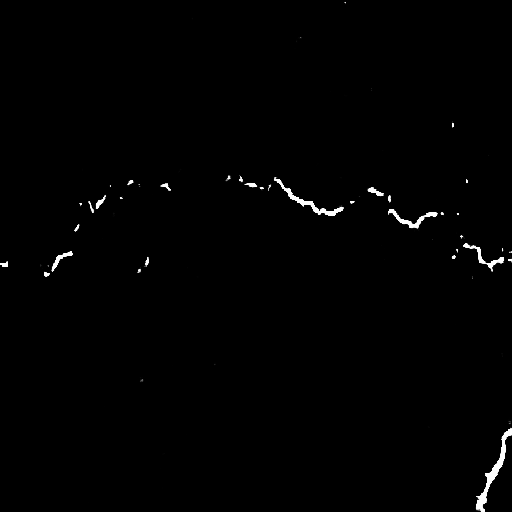}\\
    \vspace{0.1cm}
    \includegraphics[width=2.0cm,height=2.0cm]{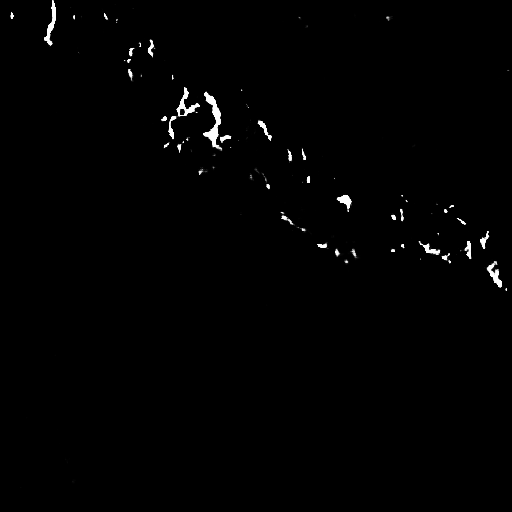}\\
    \vspace{0.1cm}
\end{minipage}\hspace{-3mm}
}%
\subfigure[Ours]{
\begin{minipage}{0.12\linewidth}
    \centering
    \includegraphics[width=2.0cm,height=2.0cm]{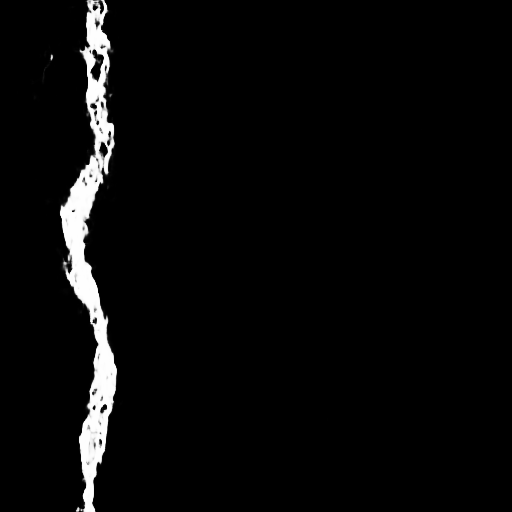}\\
    \vspace{0.1cm}
    \includegraphics[width=2.0cm,height=2.0cm]{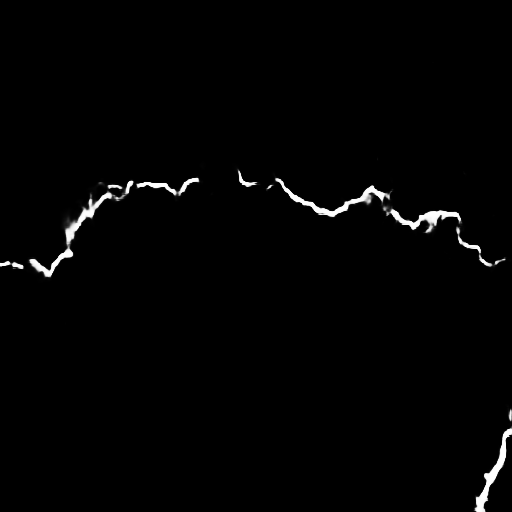}\\
    \vspace{0.1cm}
    \includegraphics[width=2.0cm,height=2.0cm]{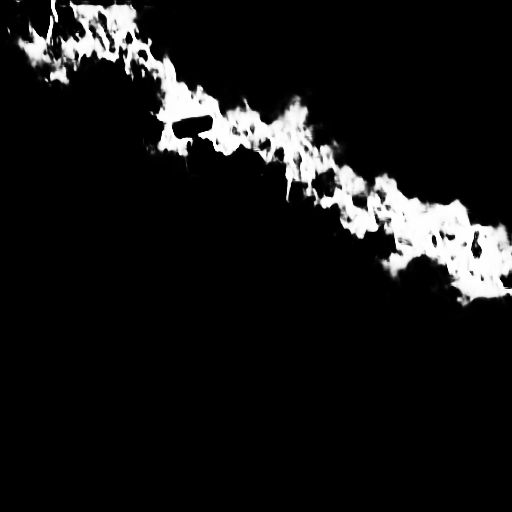}\\
    \vspace{0.1cm}
\end{minipage}\hspace{-3mm}
}%
\caption{Visualization of samples from the Crack500 dataset.}
\label{fig6_2}
\end{figure*}

\textbf{Visualization on the CFD dataset.} 
Fig. \ref{fig6} presents segmentation results on the CFD dataset under a zero-shot evaluation, comparing the proposed method to five top-performing models. The results demonstrate the superior performance of the proposed method in challenging scenarios, including thin crack patterns and faint features.

In Row 1, the proposed method accurately captures the intersecting crack network with remarkable continuity and precision, closely aligning with the ground truth. In contrast, other methods suffer from fragmentation and incomplete predictions. In Rows 2 and 4, where cracks are extremely thin and barely discernible, the proposed method successfully detects these subtle features, whereas other methods fail to identify the cracks entirely. In Row 3, the proposed method outperforms others in handling thin cracks on a textured background, delivering continuous and sharp predictions. Competing methods struggle with maintaining consistency and capturing fine details in such scenarios.

\textbf{Visualization on the Crack500 dataset.} As shown in Fig. \ref{fig6_2}, cracks in the Crack500 dataset are generally thicker with complex background textures (e.g., uneven asphalt and gravel), which differ significantly from those in the Deepcrack dataset. In zero-shot comparisons, our method achieves more accurate and complete crack extraction, demonstrating its robustness and superior generalization across diverse scenarios by effectively leveraging general prior knowledge.

These results highlight the robustness and generalization capability of the proposed method, demonstrating its effectiveness in accurately detecting thin and faint cracks across diverse and challenging conditions.

\subsection{Ablation Study}

\begin{table}
\renewcommand{\arraystretch}{1.8}
\setlength{\tabcolsep}{3.5pt}
\normalsize
% \large
\caption{Ablation experiments of the proposed modules (\%).}
\label{tabel4}
\centering
\begin{tabular}{c|ccc|ccc}
\hline
\multirow{2}{*}{Method} & \multicolumn{3}{c|}{DeepCrack} & \multicolumn{3}{c}{CFD} \\ \cline{2-7} 
                        &                               F1       & IoU      & Dice    & F1     & IoU    & Dice  \\ \hline
EdgeSAM                  & 73.79 & 59.54 & 72.61   & 51.04 & 34.88 & 49.75 \\
Unet                    & 79.64 & 67.00 & 78.41 & 37.48 & 26.45 & 37.27  \\
 \hline
Ours (Concat)                  &                    78.78 & 66.34 & 77.95  & 35.99 & 25.16 & 35.85 \\
\textbf{Ours (IGAM)}                   &                    \textbf{82.88} & \textbf{71.33} & \textbf{82.25}  & \textbf{54.48}& \textbf{39.15}& \textbf{53.83} \\ \hline
\end{tabular}
\end{table}

An ablation study is conducted to evaluate the effectiveness of FlexiCrackNet, using U-Net and EdgeSAM as baselines. 
As shown in Table \ref{tabel4}, FlexiCrackNet achieves significant improvements over U-Net, with a 3.24\% gain in F1 score, 4.33\% in IoU, and 3.84\% in Dice on the DeepCrack dataset, and a 17\% gain in F1 score, 12.7\% in IoU, and 16.56\% in Dice on the CFD dataset. Compared to EdgeSAM, our approach further improves performance, with a 9.09\% higher F1 score, 11.79\% in IoU, and 9.64\% in Dice on DeepCrack, and a 3.44\% gain in F1 score, 4.27\% in IoU, and 4.08\% in Dice on CFD. These results clearly demonstrate the superior performance of our method compared to the baseline U-Net and EdgeSAM.

To further demonstrate the effectiveness of our proposed IGAM module, we compared it with the Concat (Concatenate) operation for fusing General Prior Features and crack-specific features. As shown in Table \ref{tabel4}, the Concat operation results in performance drops on both the DeepCrack and CFD datasets compared to the U-Net baseline. This decline may be due to optimization challenges, as the General Prior Features might contain irrelevant information that misaligns with the model's feature space, especially given U-Net's limited parameter capacity. In contrast, our IGAM module effectively addresses this issue by dynamically aligning and weighting the features, ensuring only relevant information is utilized. This leads to significant performance improvements over Concat, with IGAM outperforming Concat by 4.10\% in F1 score, 4.99\% in IoU, and 4.30\% in Dice coefficient on DeepCrack, and by 18.49\% in F1 score, 13.99\% in IoU, and 17.98\% in Dice coefficient on CFD. These results validate the superiority of IGAM's fusion strategy.

\begin{figure}
\subfigure[Images]{
\begin{minipage}{0.225\linewidth}  % 
    \includegraphics[width=2.0cm,height=2.0cm]{figure/DeepCeack/1/img.jpg}\\
    \vspace{-0.33cm}
    \includegraphics[width=2.0cm,height=2.0cm]{figure/DeepCeack/3/img.jpg}\\
    \vspace{-0.33cm}
    \includegraphics[width=2.0cm,height=2.0cm]{figure/DeepCeack/2/img.jpg}\\
    \vspace{-0.33cm}
    \includegraphics[width=2.0cm,height=2.0cm]{figure/cfd/4/img.jpg}\\
    \vspace{-0.33cm}
\end{minipage} \hspace{-4mm}  % 
}
\subfigure[Ground Truth]{
\begin{minipage}{0.225\linewidth}  % 
    \includegraphics[width=2.0cm,height=2.0cm]{figure/DeepCeack/1/gt.jpg}\\
    \vspace{-0.33cm}
    \includegraphics[width=2.0cm,height=2.0cm]{figure/DeepCeack/3/gt.png}\\
    \vspace{-0.33cm}
    \includegraphics[width=2.0cm,height=2.0cm]{figure/DeepCeack/2/gt.jpg}\\
    \vspace{-0.33cm}
    \includegraphics[width=2.0cm,height=2.0cm]{figure/cfd/4/gt.jpg}\\
    \vspace{-0.33cm} 
\end{minipage}
}
\hspace{-4mm}
\subfigure[Unet]{
\begin{minipage}{0.225\linewidth}  % 
    \includegraphics[width=2.0cm,height=2.0cm]{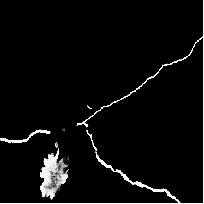}\\
    \vspace{-0.33cm}
    \includegraphics[width=2.0cm,height=2.0cm]{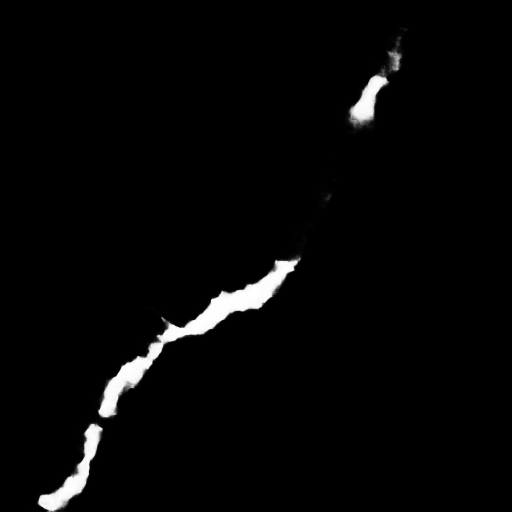}\\
    \vspace{-0.33cm}
    \includegraphics[width=2.0cm,height=2.0cm]{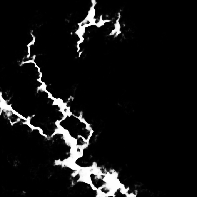}\\
    \vspace{-0.33cm}
    \includegraphics[width=2.0cm,height=2.0cm]{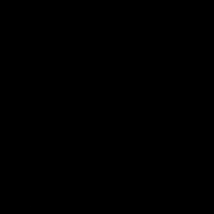}\\
    \vspace{-0.33cm}
\end{minipage} \hspace{-4mm}  % 
}
\subfigure[Ours]{
\begin{minipage}{0.225\linewidth}  % 
    \includegraphics[width=2.0cm,height=2.0cm]{figure/DeepCeack/1/ours.jpg}\\
    \vspace{-0.33cm}
    \includegraphics[width=2.0cm,height=2.0cm]{figure/DeepCeack/3/ours.jpg}\\
    \vspace{-0.33cm}
    \includegraphics[width=2.0cm,height=2.0cm]{figure/DeepCeack/2/ours.jpg}\\
    \vspace{-0.33cm}
    \includegraphics[width=2.0cm,height=2.0cm]{figure/cfd/4/ours.jpg}\\
    \vspace{-0.33cm} \hspace{-4mm}  % 
\end{minipage}
}
\caption{Comparison with the baseline.}
\label{fig7}
\end{figure}

Fig. \ref{fig7} further illustrates the effectiveness of our approach, comparing the prediction results of our model (d) with those of the baseline U-Net (c). The visual comparisons emphasize the robustness of our method across a variety of challenging scenarios, including blurred inputs, complex and noisy backgrounds, and highly textured surfaces. In all cases, our method consistently outperforms the baseline, showcasing clear advantages in segmentation accuracy.

These findings highlight the ability of FlexiCrackNet to effectively integrate additional general knowledge into the U-Net framework, resulting in substantial improvements in segmentation precision and generalization capability. The enhanced performance across multiple datasets and challenging conditions underscores the robustness and practical applicability of our approach.

\section{CONCLUSION}
In this paper, we propose FlexiCrackNet, a novel crack segmentation pipeline designed to address the limitations of traditional deep learning-based models and the ``pre-training + fine-tuning'' paradigm, particularly in resource-constrained environments. FlexiCrackNet effectively integrates advanced features extracted from large-scale pre-trained models with the scalability and adaptability required for downstream tasks. 
By fusing task-specific domain information with general knowledge in an efficient and compatible manner, the pipeline demonstrates significant improvements in segmentation performance and zero-shot generalization across diverse datasets. 
Experimental results highlight FlexiCrackNet’s consistent superiority over state-of-the-art models, particularly in handling complex, noisy crack samples, establishing it as a robust, efficient, and adaptable solution for real-world infrastructure maintenance and automated road monitoring systems.

Our future work could explore integrating real-time self-learning capabilities into FlexiCrackNet, allowing the model to continuously adapt to evolving road conditions and newly encountered defect patterns similar to cracks. Furthermore, FlexiCrackNet lays the foundation for a new paradigm in leveraging features from vision foundation models for a wide range of downstream tasks beyond crack segmentation, promising broader applicability and impact in intelligent visual perception systems.

\bibliographystyle{IEEEtran}
\bibliography{bibliography}

\vfill

\end{document}